\documentclass[journal]{IEEEtran}

\usepackage{cite}
\usepackage[cmex10]{amsmath}
\usepackage[tight,footnotesize]{subfigure}
\usepackage{caption}
\usepackage{stfloats}
\usepackage{url}
\usepackage{graphicx}
\usepackage{color}
\usepackage{xcolor}
\usepackage{placeins}
\usepackage{tabularx,colortbl}
\usepackage{ifthen}

\usepackage{cuted}
\usepackage{array}
\usepackage{makecell}
\usepackage{amssymb}
\usepackage{enumitem}
\usepackage{makecell}
\usepackage{multirow}
\usepackage{algorithm}
\usepackage{algpseudocode}

\usepackage{comment}

\newcommand{\ZP}[1]{{\color{black}#1}}

\newcommand{\ZVP}[1]{{\color{black}#1}}

\begin{document}



\title{\ZVP{Photon Splatting: A Physics-Guided Neural Surrogate for Real-Time Wireless Channel Prediction}}

\author{Ge~Cao,~\IEEEmembership{Student Member,~IEEE,}
        Gabriele~Gradoni,~\IEEEmembership{Member,~IEEE,}
        and~Zhen~Peng,~\IEEEmembership{Senior~Member,~IEEE}

\thanks{Ge Cao and Z. Peng are with the Center for Computational Electromagnetics, Department of Electrical and Computer Engineering, The Grainger College of Engineering, University of Illinois at Urbana-Champaign, Urbana, IL 61801 USA (e-mail: gecao2@illinois.edu;  zvpeng@illinois.edu).}
\thanks{G. Gradoni is with the Institute for Communication Systems, University of Surrey, Guildford, Surrey, UK, and he is also an adjunct professor at the Department of Electrical and Computer Engineering, University of Illinois at Urbana-Champaign, USA (e-mail: g.gradoni@surrey.ac.uk)}
}


\maketitle

\begin{abstract}

\ZVP{We present Photon Splatting, a physics-guided neural surrogate model for real-time wireless channel prediction in complex environments. The proposed framework introduces surface-attached virtual sources, referred to as photons, which carry directional wave signatures informed by the scene geometry and transmitter configuration. At runtime, channel impulse responses (CIRs) are predicted by splatting these photons onto the angular domain of the receiver using a geodesic rasterizer. The model is trained to learn a physically grounded representation that maps transmitter-receiver configurations to full channel responses. Once trained, it generalizes to new transmitter positions, antenna beam patterns, and mobile receivers without requiring model retraining. We demonstrate the effectiveness of the framework through a series of experiments, from canonical 3D scenes to a complex indoor caf\'e with 1,000 receivers. Results show 30 millisecond-level inference latency and accurate CIR predictions across a wide range of configurations. The approach supports real-time adaptability and interpretability, making it a promising candidate for wireless digital twin platforms and future 6G network planning.
}
\end{abstract}


\section{Introduction}
\subsection{Research Motivation}

\ZVP{
Wireless channel prediction plays a pivotal role in the design, optimization, and operation of next-generation wireless systems. As networks evolve toward higher frequencies, denser deployments, and increased complexity, there is a growing need for site-specific channel models that capture how electromagnetic (EM) waves interact with the actual physical environment. This need is especially critical in the context of wireless digital twins (WDTs). They are emerging platforms that create synchronized, predictive models of the wireless environment to support applications such as autonomous driving, Industry 4.0, extended reality, and immersive multimedia experiences \cite{DigitalTwin2022}. The vision is to create real-time virtual replicas of wireless systems that enable predictive analysis, performance optimization, and user-centric adaptation with ultra-low latency, high reliability, and stringent quality-of-service. Currently, WDTs have become an important focus of 6G research and standardization activities.

Despite their promise, several core challenges need be addressed to realize WDTs at scale. First, high-fidelity physical-layer modeling is essential to capture complex radio wave interactions with propagation environments. Second, real-time channel prediction is crucial to enable timely system responses such as beamforming optimization,  waveform design, and resource allocation, based on predicted channel states. 
Third, dynamic adaptability and relightability are required so that the modeling framework can seamlessly handle changes in user deployments and antenna configurations without the need for retraining or scene recomputation. Achieving these objectives simultaneously remains a fundamental challenge for practical physical-layer WDT deployment. 
}

\subsection{Limitations of Existing Methods}

\ZVP{Ray-tracing methods have long been used for wireless propagation modeling in both indoor and outdoor environments \cite{RadioRayTracing}. While they provide physical fidelity by simulating wave-environment interactions such as reflection, diffraction, and scattering, their computational cost grows significantly with scene complexity and the number of multipath interactions. This makes ray tracing impractical for real-time applications where rapid and continuous updates are required.

Recent advances in neural network-based methods have opened new avenues for accelerating radio propagation modeling \cite{Ray-Launching-Neural-2014, 8740286, DeepRay2022, Costas_2022, Seretis_2022_CNN, Seretis_2023, Liu2023_GNN, lee2023pmnet}. These approaches have demonstrated promise in predicting path loss and other large-scale parameters using 2D inputs like floor plans or heatmaps. Most of them operate under simplified assumptions and struggle to generalize to full 3D environments with arbitrary user deployments. Moreover, they typically do not predict time-resolved channel impulse responses (CIRs) \cite{tse2005fundamentals}, which are essential for physical-layer simulation, waveform optimization, and protocol design. Finally, none of these approaches achieves real-time performance, which is defined as the ability to make predictions within millisecond-level latency, a measure of the system's responsiveness to changing radio conditions. 

As one of the state-of-the-art industrial solutions, NVIDIA's Aerial Omniverse Digital Twin (AODT) platform integrates high-performance ray tracing with AI to support wireless system design. AODT offers a system-level simulation environment that enables physically accurate modeling of 5G and 6G networks.
While AODT represents a significant advancement in system integration, it still encounters challenges in computational scalability and dynamic adaptability, making it less suited for real-time WDT operations where channel predictions must be continuously updated in response to changing user locations, antenna configurations, and environmental conditions.
}

\subsection{Proposed Approach: Photon Splatting}

\ZVP{Given the need for real-time, adaptive, and physically grounded channel prediction, we propose Photon Splatting, a physics-guided neural surrogate framework that unifies electromagnetic modeling with neural operator learning. Instead of tracing individual ray paths, Photon Splatting models radio propagation through a set of surface-attached virtual sources, referred to as photons. Each photon carries a learned wave signature, a compact, physically meaningful representation of how waves scatter and radiate from its position, conditioned on the transmitter's configuration.  

To efficiently handle the resulting spatial and angular dependencies, the framework employs Spherical Harmonics for encoding directional properties and Fourier Neural Operators (FNOs) \cite{FNO2020} to model the transmitter-photon interaction. This design enables the model to capture multipath accumulation and adapt to new transmitter or receiver configurations without retraining. At runtime, directional contributions from the photons are projected onto the receiver's angular domain using a spherical rasterizer. This allows the channel impulse response to be reconstructed in real time, without launching rays or recomputing wave trajectories.

While conceptually related to the field of neural rendering, Photon Splatting is grounded in EM wave physics and tailored for wireless communication. It supports interpretability, physical consistency, and real-time adaptability, making it well-suited for emerging wireless digital twin applications.

The remainder of the paper is organized as follows: Section II reviews related work on physics-based and neural propagation modeling. Section III introduces the Photon Splatting framework and its physical foundations, neural modeling components, and training strategies. Section IV presents numerical experiments and performance analysis. Section V concludes the paper and discusses limitations and future directions.}

\section{Background and Related Work}

\ZVP{In this section, we review existing and related work for radio propagation and neural rendering, organized along three themes: traditional physics-based techniques, neural rendering methods from computer graphics, and recent neural surrogate models for wireless applications. This organization reflects the conceptual foundation of our work, which integrates electromagnetic theory with insights from neural rendering to enable efficient, physically grounded channel prediction.
}

\subsection{Traditional Modeling Techniques}

\ZVP{\subsubsection{Ray Tracing}
Due to the high computational cost of full-wave numerical solvers \cite{865237, Tsang_2004, 1504967, Sevgi_2007_Hybrid,  Sarris_FDTD_Fading, 8485766}, ray tracing has become a popular approximation technique for high-frequency wireless propagation modeling in complex indoor and urban environments \cite{Aguado_2000, Sarkar_Ray_2001, 901882, Yun2015, Raytracing_tutorial_2019}. By leveraging geometric optics, ray tracing models key interactions like reflection, diffraction, and scattering through deterministic ray paths. In parallel, ray tracing has played a foundational role in computer graphics since the introduction of the rendering equation \cite{RenderEquation1986}, enabling photorealistic image synthesis. Recent advances such as gradient-domain path tracing \cite{gdpt2015} and path guiding \cite{muller2017practical} have improved efficiency and convergence speed of rendering pipelines.

\subsubsection{Photon Mapping}
Photon mapping offers an alternative to traditional ray tracing by simulating light transport through a particle-based approach \cite{PhotonMapping1996, ProgressivePhotonMapping2008, StochasticProgressivePhotonMapping2009}. Rather than tracing rays from the camera, photons are emitted from the light source, interact with surfaces, and deposit their energy onto a global photon map. Camera rays then gather nearby photons to estimate radiance, yielding a spatially smooth and variance-reduced radiance field. 
An illustration of the photon mapping algorithm in rendering is provided in Fig. \ref{fig: photon mapping}.

Photon Mapping concept presents an appealing analogy for modeling wave propagation: photons serve as discrete carriers of energy, and their interactions encode local scattering information. While developed for rendering, this particle-based view of energy transport aligns well with EM energy flow, offering a conceptual inspiration for the approach we propose in this work.

\begin{figure}[!ht]
\centering
\includegraphics[width=0.8\linewidth]{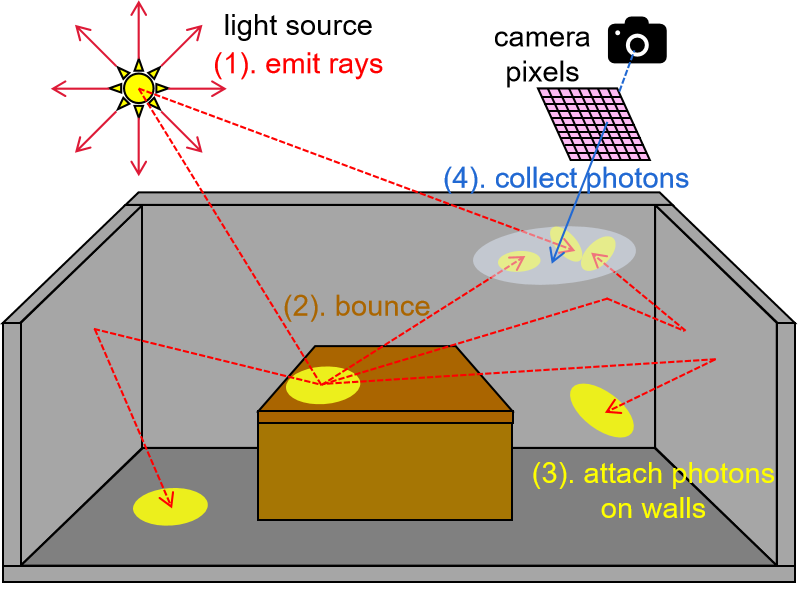}
\caption{Illustration of Photon Mapping. (1) Photons are emitted from light sources and (2) bounce through the scene, (3) when conditions are met, photons are deposited on surfaces, (4) camera pixels collect nearby photons to estimate radiance.
}
\label{fig: photon mapping}
\end{figure}
}

\subsection{Neural Rendering}

\ZVP{Neural rendering has advanced rapidly following the introduction of Neural Radiance Fields (NeRF) by Mildenhall et al. \cite{NeRF}, which models the radiance field as a continuous function parameterized by a neural network. NeRF enables novel view synthesis and high-fidelity 3D scene reconstruction from sparse images. Since its debut, numerous extensions have sought to improve inference speed, relighting capabilities, and visual quality \cite{Renerf2023, EyeNeRF2022}. However, the computational cost of NeRF inference remains high, posing challenges for real-time applications such as gaming, augmented reality, or virtual reality.

More recently, 3D Gaussian Splatting has emerged as a promising technique for real-time neural rendering \cite{3DGS}. By representing the scene with anisotropic Gaussians centered on point clouds, these methods achieve high rendering quality with millisecond-level latency. Variants such as 2D Gaussian Splatting and MiniGS further reduce memory usage and computational complexity, offering lightweight yet effective pipelines for real-time view synthesis \cite{2DGS, fang2024mini}.

These advances offer valuable algorithmic tools for modeling spatially distributed energy transport, a desirable property for wireless channel modeling. However, directly adapting these techniques to radio frequency (RF) propagation involves nontrivial complexities, such as antenna radiation patterns, wave polarization, and vectorial EM fields, which go beyond scalar light transport. As such, realizing physically meaningful RF rendering requires deeper integration with the underlying physics of wave propagation.
}



\begin{table*}[!t]
    \centering
    \caption{Comparison of Neural Propagation Models. 
    Key capabilities include relightability (generalizing to new Tx/Rx positions), dynamic adaptability (handling unseen beam patterns), and scalability to complex 3D scenes.}
    \label{tab:comparison with previous}
    \begin{tabular}{|c|c|c|c|c|c|c|}
        \hline
        \textbf{Method} & \textbf{Real-Time ($\sim 10^3$ Rx)} & \textbf{3D Modeling} & \textbf{Relightability} & \textbf{Dynamic Adaptability} & \textbf{CIR Prediction} & \textbf{Scalability} \\
   	 \hline
        [3]-[10] & $\times$ & $\times$ & $\times$ & $\times$ & $\times$  & -\\
        \hline
        WiNert & $\times$ & $\checkmark$ & $\checkmark$ & $\times$ & $\checkmark$  & $\times$\\
        \hline
        RayProNet & $\checkmark$ & $\checkmark$ & $\checkmark$ & $\times$ & $\times$ & $\checkmark$ \\
        \hline
        RF-3DGS & $\times$ & $\checkmark$ & $\times$ & $\times$ & $\checkmark$ & $\checkmark$ \\
        \hline
        WRF-GS & $\times$ & $\checkmark$ & $\times$ & $\times$ & $\times$ & $\checkmark$ \\
        \hline
        \textbf{Photon Splatting (Ours)} & $\checkmark$ & $\checkmark$ & $\checkmark$ & $\checkmark$ & $\checkmark$ & $\checkmark$\\
        \hline
    \end{tabular}
\end{table*}

\subsection{Neural Wireless Propagation}
\ZVP{Neural surrogate models for radio propagation have gained significant interest as faster alternatives to traditional physics-based simulations. Early works often focused on 2D inputs, such as floorplans or top-down signal heatmaps, to learn large-scale path loss or signal coverage distributions \cite{Ray-Launching-Neural-2014, 8740286, DeepRay2022, Costas_2022, Seretis_2022_CNN, Seretis_2023, Liu2023_GNN, lee2023pmnet}. While these models are computationally efficient, they offer limited physical interpretability and struggle to generalize to complex 3D environments. In particular, they fall short in capturing key wave phenomena such as multipath propagation and angular (directional) structure.

Motivated by recent progress in neural rendering, several recent efforts have extended surrogate models into 3D. WiNERT adopts a NeRF-style encoder to reconstruct spatially continuous channel information from sparse measurements \cite{WiNert2023}. RayProNet employed neural point light fields to estimate path loss across irregular 3D indoor layouts \cite{raypronet}, offering improved scene generalization but limited to scalar field outputs. {RF-3DGS} proposed a 3D Gaussian splatting model that reconstructs a Radio Radiance Field (RRF), learning spherical harmonic profiles for each Gaussian to render spatial-CSI images at fixed receiver viewpoints \cite{RF3DGS2024}. {WRF-GS} adapted Gaussian splatting for real-time wireless radiation field reconstruction across angular receivers \cite{WRFGS2024}.

Despite these advances, several limitations remain. Most existing models assume fixed transmitter or receiver positions after training and lack mechanisms for supporting dynamic reconfiguration or adaptability to unseen antenna patterns. Additionally, they do not directly predict time-domain channel impulse responses and offer limited integration with physical-layer features such as beamforming vectors or vectorial field characteristics. 
A comparative summary of these approaches, along with our proposed Photon Splatting framework, is provided in Table~\ref{tab:comparison with previous}.
}

\section{Methodology}



\subsection{Problem Statement and Notation}

\ZP{
We consider the task of modeling the wireless propagation channel between a pair of transmitter (Tx) and receiver (Rx) in a complex physical environment. The proposed framework retains the same functional inputs/outputs as a ray-tracing-based propagation model. Specifically, it takes as input: (1) the geometric representation of the scene (e.g., triangle mesh or point cloud), and (2) the Tx and Rx configurations, including their spatial positions and antenna patterns. The output is the channel response between any Tx-Rx pair in the environment. One important quantity of interest is the channel frequency response (CFR), defined as:
\begin{align}
H(f) \!=\!\! \sum_{n=1}^{N} \frac{\lambda}{4\pi} \mathbf{C}_\mathrm{R}^\mathrm{H}(\theta_n^\mathrm{rx}, \phi_n^\mathrm{rx}) \!\cdot\! {\mathbf{T}}_n e^{-j2\pi f \tau_n} \!\cdot\! \mathbf{C}_\mathrm{T}(\theta_n^\mathrm{tx}, \phi_n^\mathrm{tx}), \!
\label{Equation: CFR}
\end{align}
where: $N$ is the number of propagation paths, $\lambda$ is the wavelength, $\mathbf{C}_\mathrm{T}, \mathbf{C}_\mathrm{R} \in \mathbb{C}^{2} $ are antenna field patterns at the Tx and Rx,
$(\theta_n^\mathrm{tx}, \phi_n^\mathrm{tx})$ and $(\theta_n^\mathrm{rx}, \phi_n^\mathrm{rx})$ are the departure and arrival angles. ${\mathbf{T}}_n \in \mathbb{C}^{2 \times 2} $ is the transfer function matrix that characterizes the transformation of the EM radiated field from the Tx to the Rx along the $n$-th path \cite{aoudia2025sionna}. It captures the cumulative effects of reflections, diffractions, and interactions with scene surfaces. 
$\tau_n$ is the total delay of the $n$-th path.

Taking the inverse Fourier transform, the corresponding baseband channel impulse response (CIR) is \cite{tse2005fundamentals}:
\begin{align}
h_b(\tau) = \sum_{n=1}^{N} a_n e^{-j2\pi f_c \tau_n} \delta(\tau - \tau_n),
\label{Equation: CIR}
\end{align}
where \( a_n \in \mathbb{C} \) encodes path amplitude and phase, $f_c$ is the carrier frequency.

Our objective is to develop a neural surrogate model to accelerate predicting wireless channel responses in a given environment. Once trained on a sparse set of simulations or measurements, the model generalizes to unseen Tx and Rx configurations, including new antenna patterns and placements, without retraining. The learning task is highly nontrivial due to the spatial, angular, and temporal complexity of EM propagation. The success of the model depends on constructing a physically grounded, data-efficient representation that reflects the structure of Maxwellian interactions while remaining amenable to learning.
}

\subsection{Physical Foundation of Photon Splatting}

\ZP{The Photon Splatting framework is built on the idea of replacing explicit ray-path tracing with a set of surface-attached virtual sources, which we refer to as photons. These photons are not rays or particles in the classical sense, but rather localized abstractions of EM wave interaction states. Each photon represents a re-radiating element embedded on the scene surface, encoding wave interaction signatures with the geometry.

Conceptually, we can view each photon as a virtual antenna that emits toward the receiver. Angular wave information is encoded using Spherical Harmonics defined over a geodesic sphere attached to the photon location. This allows each photon to capture direction-dependent propagation effects and enables spatial/angular interpolation at the receiver. 

\ZVP{Specifically, each photon is associated with a set of physical and learned attributes: a spatial location \( \mathbf{x}_i \in \mathbb{R}^3 \), Tx radiation direction \( (\theta_i^\mathrm{tx}, \phi_i^\mathrm{tx}) \), an accumulated path delay \( \tau_i \), a complex-valued transfer function matrix \( \mathbf{T}_i \in \mathbb{C}^{2 \times 2} \) encoding wave-surface interaction, and a set of Spherical Harmonics coefficients that encode the photon's angular re-radiation behavior.}

\ZVP{Spherical Harmonics are well-suited for this task, as they offer a compact and rotation-aware representation of angular variation. In our framework, they dynamically modulate each photon's directional response based on the relative angle to the receiver, enabling smooth and physically meaningful interpolation of wave contributions as the receiver moves throughout the environment. An illustration of spherical harmonics basis functions is shown in Fig.~\ref{fig: Spherical Harmonics}.
}


\begin{figure}[!ht]
\centering
\includegraphics[width=\linewidth]{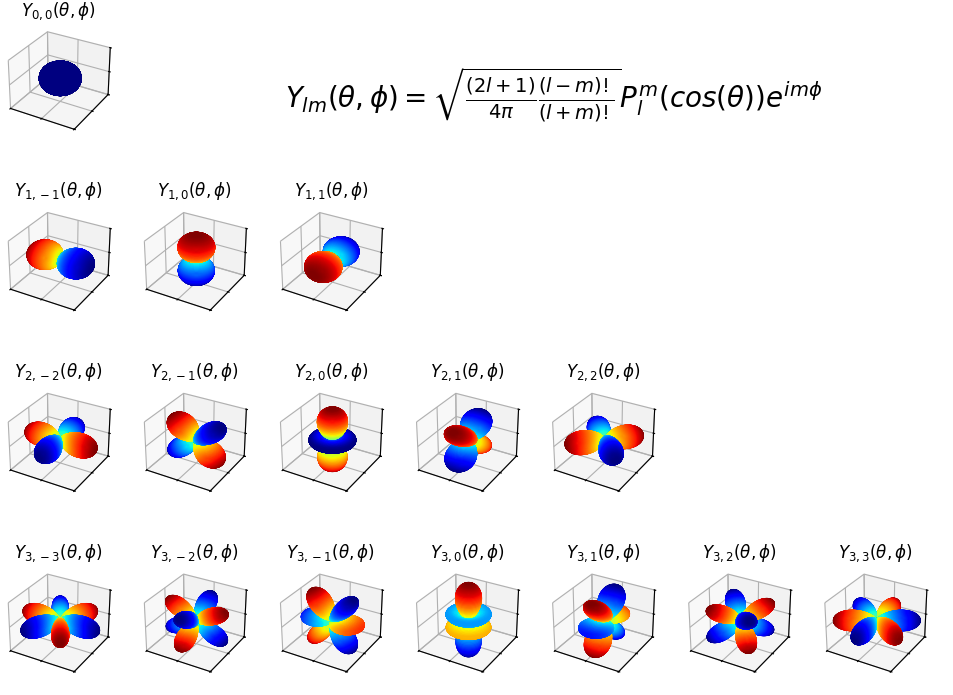}
\caption{
\ZVP{
Illustration of Spherical Harmonics basis functions at order $l=3$, resulting in a total of $(l + 1)^2 = 16$ basis components. Higher-order expansions are employed in practice to capture high-frequency directional information accurately.
}
}
\label{fig: Spherical Harmonics}
\end{figure}

At runtime, a receiver queries directional contributions from all photons in the scene. Each photon contributes a ``splat'' to the angular domain, filtered by the receiver's orientation and antenna pattern. The angular splatting operation accumulates these multipath contributions over a spherical rasterizer, forming the basis for reconstructing the channel impulse response. This process eliminates the need for ray launching or full-scale simulation during inference.
}

\subsection{Channel Computation via Photon Splatting}

\ZP{
To model the directional wave contributions from surface photons to a receiver, we discretize the angular domain using a geodesic spherical rasterizer. Each photon is equipped with a local angular basis, and its directional emission profile is encoded using a nearest-pixel scheme. This forms the foundation for the angular splatting process.

When queried by a receiver, each photon contributes a directional contribution, referred to as a splat toward the receiver's angle of arrival \( (\theta^\mathrm{rx}, \phi^\mathrm{rx}) \). The magnitude and direction of this splat are determined by the alignment between the receiver's antenna pattern and the angular emission pattern encoded by the photon. This operation can be viewed as applying a weighted angular filter (the antenna pattern) over a spatial distribution of photon-originated emissions.

To represent the contribution of each photon in a compact and learnable form, we define a \textit{wave signature vector}. This vector encapsulates key propagation parameters, such as path delay, transmission direction, and scattering behavior, in a fixed-size format suitable for neural prediction and efficient angular projection. It serves as the fundamental representation learned by the neural surrogate and evaluated during splatting.

The wave signature vector with the $i$-th photon is defined as:
\begin{align}
\mathbf{s}_i = \left[ \frac{1}{td_i}, \theta_i^\mathrm{tx}, \phi_i^\mathrm{tx}, T_i^{11}, T_i^{12}, T_i^{21}, T_i^{22} \right],
\end{align}
where $td_i$ is the total path length (proportional to delay), $ (\theta_i^\mathrm{tx}, \phi_i^\mathrm{tx})$ are the transmission angles, and $\mathbf{T}_i \in \mathbb{C}^{2 \times 2}$ is the local transfer matrix representing wave-surface interaction at the photon's location.

The contribution of photon $i$ to the overall channel response is computed as:
\begin{align}
\begin{cases}
\tau_i = \frac{d_0 + td_i}{c}, \\
a_i = \frac{\lambda}{4 \pi} \mathbf{C}_\mathrm{R}^\mathrm{H}(\theta_i^\mathrm{rx}, \phi_i^\mathrm{rx}) \cdot \mathbf{T}_i  \cdot \mathbf{C}_\mathrm{T}(\theta_i^\mathrm{tx}, \phi_i^\mathrm{tx}),  
\end{cases}
\end{align}
where \( \mathbf{C}_\mathrm{T} \) and \( \mathbf{C}_\mathrm{R} \) are the complex-valued antenna patterns at the Tx and Rx, $d_{0}$ is the distance between Rx and the visible photon, and \( c \) is the speed of light. These values are inserted into Eq.~\eqref{Equation: CIR} to construct the CIR.

This computation is repeated across all relevant photons. Each photon's directional contribution is projected onto the receiver's local angular basis through a process we refer to as \textit{spherical rasterizer splatting}, as illustrated in Fig.~\ref{fig:rasterizer_splatter}. 

\begin{figure}[ht]
\centering
\includegraphics[width=0.48\textwidth]{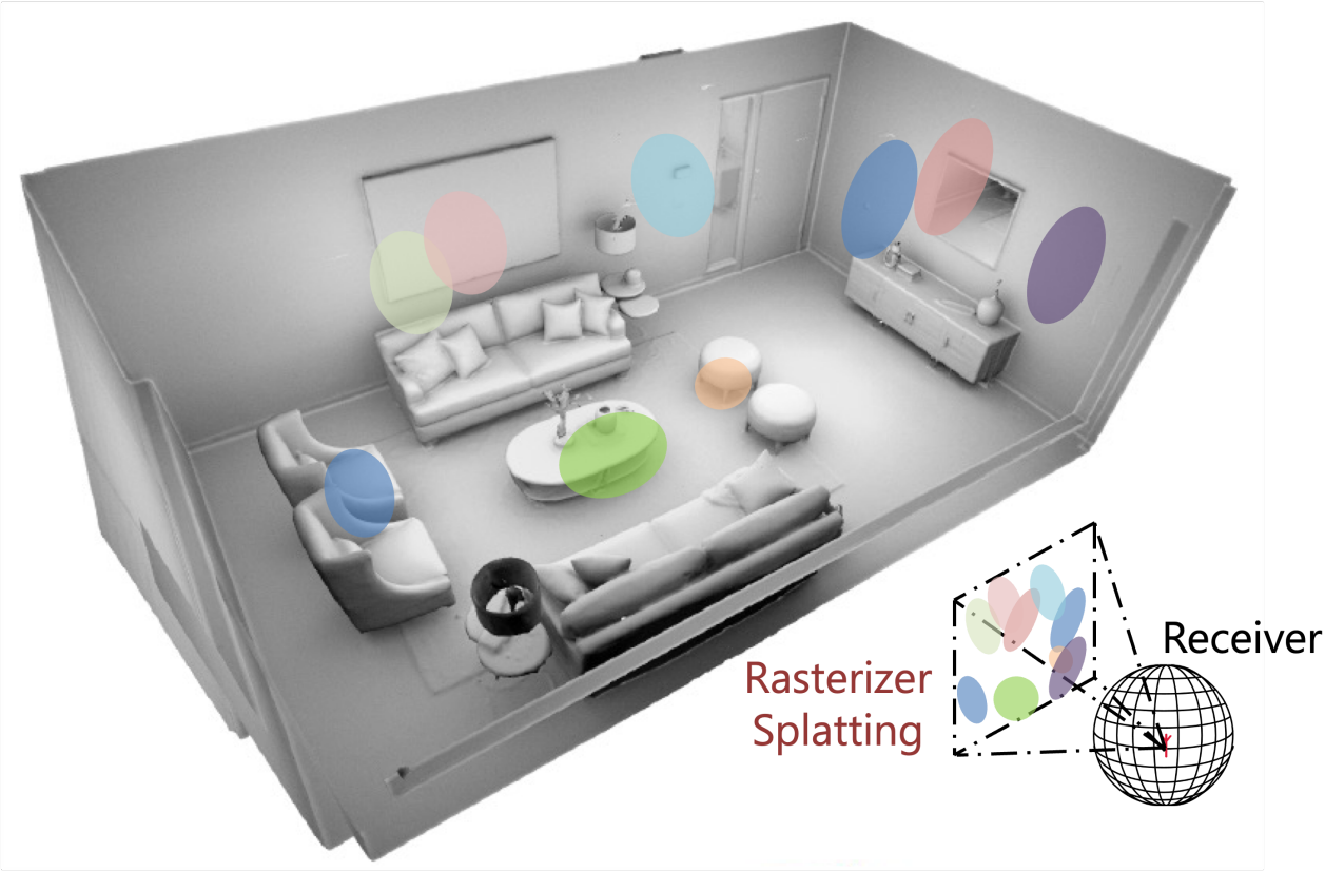}
\caption{Illustration of spherical rasterizer splatting. Each surface-attached photon contributes a directional splat onto the receiver's geodesic rasterizer, enabling angularly resolved channel accumulation.}
\label{fig:rasterizer_splatter}
\end{figure}

This angular splatting formulation enables efficient accumulation of multipath contributions while preserving spatial and directional resolution. Because the photon signatures are precomputed and splatted only onto a sparse set of angular bins, the model achieves both accuracy and real-time performance. The photon wave signatures \( \mathbf{s}_i \), which govern each splat's amplitude, directionality, and delay, are predicted by a neural surrogate model described in the next section.
}

\subsection{Neural Surrogate Architecture}
\ZP{As alluded earlier, each surface-attached photon contributes to the channel response through a directional splat, governed by a wave signature that encodes its temporal delay, angular emission, and radiation behavior. In this subsection, we describe how these wave signatures are learned using a lightweight, physics-guided neural model. Rather than learning the full channel response, the model focuses on predicting compact, interpretable features per photon, thereby enabling efficient and generalizable channel computation through angular splatting.
}

\ZP{The proposed neural encoder takes as input the photon positions \( \mathbf{p}_\mathrm{ph} \in \mathbb{R}^{n_\mathrm{ph} \times 3} \) and transmitter positions \( \mathbf{p}_\mathrm{tx} \in \mathbb{R}^{n_\mathrm{tx} \times 3} \), where \( n_\mathrm{ph} \) and \( n_\mathrm{tx} \) denote the number of photons and transmitters, respectively. This spatial configuration encodes the relative geometry and launch conditions that govern the EM field interactions.

The model produces a set of wave signatures \( \mathbf{W} \in \mathbb{C}^{n_\mathrm{ph} \times n_\mathrm{bins} \times 7} \), where each 7-dimensional vector encodes: the inverse path length (related to delay), the departure angles \( (\theta, \phi) \), and the four complex components of the \( 2\times2 \) transfer matrix \( \mathbf{T}_i \). Each bin corresponds to a discrete direction in the geodesic angular rasterizer used during photon splatting.

To support fast inference, the framework integrates efficient wave aggregation through splatting with the power of the Fourier Neural Operator (FNO) \cite{FNO2020}. FNO transforms input data from the physical domain into the frequency domain, applies a linear operator to capture spatial-spectral patterns, and decodes the processed data back into the physical domain. This transformation allows the network to handle high-dimensional spatial data efficiently, capturing complex interactions such as reflection, diffraction, and scattering. The neural network architecture consists of an MLP encoder for photon and transmitter positions, followed by four FNO layers, and concludes with a two-layer MLP decoder.

The choice of FNO is physically motivated. In classical antenna theory, the far-field radiation pattern is recovered via a spatial Fourier transform of current distributions. By analogy, the spatial distribution of surface-attached photons defines a structured radiative source whose angular behavior is naturally captured in the spectral domain. FNO layers thus provide a principled and scalable way to learn global angular patterns, while the MLP decoder reconstructs fine-grained directional details at each photon location.
}


The network is trained using a composite loss that penalizes errors in both amplitude and delay:
\begin{align}
\mathcal{L} = \frac{1}{N_l} \sum_{i=1}^{N_l} \left| a_i - a_i^{\mathrm{GT}} \right|^2 + \alpha \frac{1}{N_l} \sum_{i=1}^{N_l} \left| \frac{1}{td_i} - \frac{1}{td_i^{\mathrm{GT}}} \right|^2,
\label{Equation: loss function}
\end{align}
where \( a_i^\mathrm{GT} \) and \( td_i^\mathrm{GT} \) are the ground-truth channel gain and traveled distance, respectively, and \( \alpha \) is a weighting factor. The total number of training points is \( N_l = n_\mathrm{rx} \cdot n_\theta \cdot n_\phi \), where \( (n_\theta, n_\phi) \) define the angular resolution of the spherical rasterizer.


\subsection{Operational Overview of Photon Splatting}
\begin{figure*}[!ht]
\centering
\includegraphics[width=\linewidth]{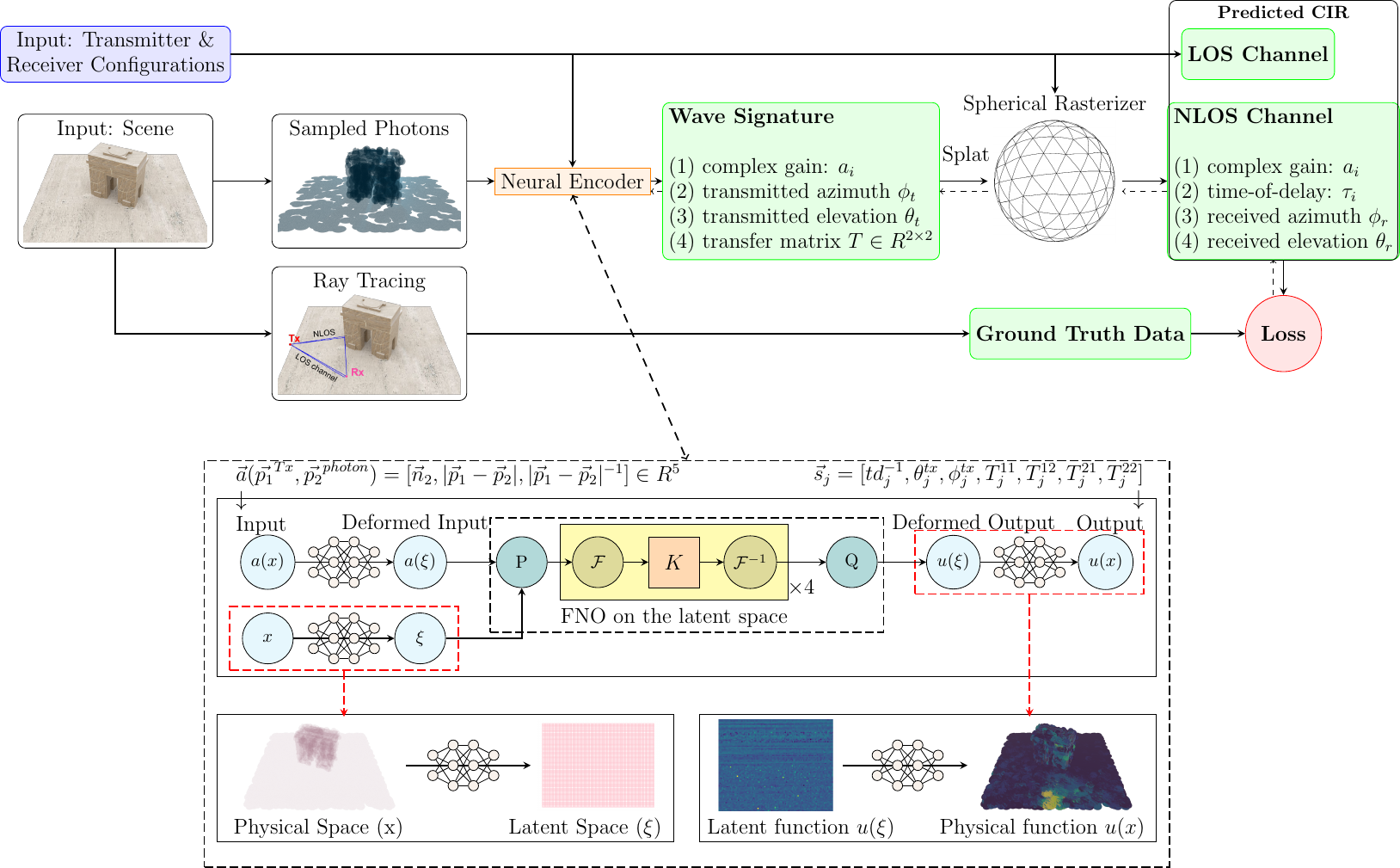}
\caption{Overview of the Photon Splatting framework. Surface-attached photons are constructed from scene geometry and learned via a neural model. At runtime, the system predicts wave signatures and aggregates angular contributions through spherical splatting to compute the CIR.
}
\label{fig: pipeline}
\end{figure*}

\ZP{
An overview of the end-to-end pipeline is shown in Fig.~\ref{fig: pipeline}. In the training phase, surface-attached photons are constructed from the scene geometry (e.g., triangle meshes or point clouds), and the neural model learns to associate each photon with a directional wave signature. This signature encodes the photon's contribution to the CIR in terms of path delay, angular emission, and scattering behavior.

At inference time, the framework operates as follows: the transmitter parameters (location and beam pattern) are projected into the angular domain and combined with the learned photon signatures. Each photon's contribution to the CIR is then evaluated through spherical rasterizer splatting, which accounts for the receiver's position and antenna pattern. The final CIR is assembled by summing all directional contributions, with gain and delay derived from the neural outputs.

It is important to note that photon positions remain fixed after preprocessing and do not change with transmitter movement. As a result, changes in transmitter location or beam pattern are captured by updating the wave signatures of fixed photons, rather than recomputing propagation paths or interaction points. This approach stands in contrast to traditional ray-tracing methods, which must recompute multipath trajectories for every new configuration, and to many neural surrogates, which require retraining to adapt to unseen array geometries. By decoupling channel prediction from path recomputation and scene regeneration, Photon Splatting offers a scalable, explainable, and physically grounded foundation for real-time wireless digital twin systems.
}

\subsection{Comparison with Radiance-Based Neural Models}
\ZP{In recent years, we have seen growing interest within the wireless communication community in adapting neural rendering techniques - originally developed in computer graphics - to wireless propagation modeling. Methods such as WiNERT, RF-3DGS, and WRF-GS adapt frameworks like Neural Radiance Fields (NeRF) or 3D Gaussian splatting to synthesize wireless channels by drawing analogies between RF signals and visual radiance. These methods provide valuable perspectives and open new directions for geometry-aware channel learning.


At the same time, wireless channel modeling at the physical layer poses unique challenges that differ from view-based rendering in computer graphics. In particular, multipath propagation, characterized by reflections, diffractions, and scattering, plays a central role in determining channel behavior. Many radiance-based approaches are optimized for spatial field reconstruction or viewpoint rendering, rather than wireless channel prediction. As a result, they often do not produce channel impulse responses and may rely on simplified or fixed antenna models. Their volumetric representations can also limit integration with physically grounded entities such as scattering matrices and vector field antenna patterns.


In contrast, Photon Splatting is explicitly designed for physically interpretable wireless channel modeling. It predicts multipath propagation and time-resolved CIRs directly, supports arbitrary transmitter and receiver configurations, and generalizes across new deployments without retraining. 
The framework uses surface-attached photons rather than volumetric Gaussians to represent wave-object interactions, enabling modularity, explainability, and efficient angular accumulation through spherical rasterization.



Finally, it is noted that in this work, we use an open-source ray-tracing engine (Sionna) \cite{sionna} to generate training and evaluation data. The framework itself is not tied to ray tracing. Photons can be generated directly from coarse geometry (e.g., point clouds) without material labels, and the model can be trained on real-world measurements. This makes the approach both flexible and practically deployable in environments where traditional ray-based modeling is infeasible.
}

\section{Numerical Experiments}

\ZP{To demonstrate the accuracy, generalization, and real-time performance of the proposed Photon Splatting framework, we conduct a suite of experiments across progressively complex wireless environments. These experiments are designed not only to validate the core technical claims of the model, but also to highlight its potential for real-world deployment in next-generation wireless systems.

We evaluate the proposed framework through three representative 3D environments, each chosen to validate a specific capability of the framework:

\begin{enumerate}
    \item \textbf{Single Building Dataset (Section IV.B):}  
    This experiment features a solitary building on an open ground and serves as a clean benchmark to validate three core capabilities: (i) full channel state prediction, including amplitude, angles, and path delay; (ii) relightability, demonstrating generalization to new transmitter locations without retraining; (iii) real-time performance, achieving 30 ms per query on a single GPU for a 900-receiver deployment. As an application use case, we demonstrate multi-user MIMO with 20 UAV-mounted transmitters and 10 mobile users, using the predicted channel transfer functions to compute zero-forcing precoding matrices.

    \item \textbf{Bistro Dataset (Section IV.C):}  This experiment involves a photorealistic caf\'e with detailed architectural and furnishing elements to test the model's performance in dense, cluttered indoor environments.
 We evaluate the model's generalization to previously unseen antenna patterns without retraining. It is also used to compare different neural encoders for an ablation study and different neural surrogates for comparison.

    \item \textbf{Wi3Rooms Dataset (Section IV.D):}  This experiment features an indoor floor plan with three rooms, drawn from the dataset of RPLAN \cite{RPLAN2019}. 
The experiment validates the model's ability to predict location-specific angular and delay characteristics of the wireless channel. As an application, we simulate wave-guided robotic navigation using only predicted CIRs. The robot leverages directional delay information from Photon Splatting to dynamically plan its path through the environment, showcasing a new capability for real-time trajectory control in wireless-aware autonomous systems.    

\end{enumerate}

A high-level summary of these experiments, their goals, and real-world relevance is provided in Table~\ref{tab:experiment_summary}.
}

\begin{table*}[t]
\centering
\caption{Summary of experiments and their application relevance.}
\label{tab:experiment_summary}
\begin{tabular}{|c|c|c|c|c|}
\hline
\textbf{Exp.} & \textbf{Scene} & \textbf{Objective} & \textbf{Application} & \textbf{Evaluation Focus} \\
\hline
IV.B & Single Building & Multiuser MIMO precoding with UAVs & Urban aerial access network & Channel matrix and ZF precoding \\
\hline
\multirow{2}{*}{IV.C} & \multirow{2}{*}{Bistro} & Scalability in large scenes & Wireless digital twin & Latency, coverage, dense Rx prediction \\
\cline{3-5}
& & Ablation and speed analysis & System design tradeoff & Runtime and component importance \\
\hline
IV.D & Indoor Room & Mobility generalization & Autonomous robot path planning & CIR accuracy over trajectory \\
\hline
\end{tabular}
\end{table*}

\subsection{Data Preparation and Hyperparameters}
\ZP{The proposed Photon Splatting pipeline is implemented using Pytorch for neural network training \cite{PyTorch}, CUDA for GPU-accelerated differentiable rasterizer, and Taichi for interactive visualization and GUI \cite{hu2019taichi}.  All datasets are generated, trained, and tested using an NVIDIA RTX A6000. 
}

\ZP{Each dataset consists of three main components: (1) the surface geometry of the environment; (2) transmitter and receiver configurations, including 3D positions and antenna radiation patterns; and (3) ray-traced ground truth (GT) data generated using Sionna. Specifically, for each transmitter-receiver pair, the ray tracer produces a set of multipath components indexed by $n = 1,\dots,N$. The GT data in each propagation path includes the time-of-delay $\tau_n$, arrival angles, and a complex channel gain ${a}_n$.
}

\ZP{Table~\ref{table: hyperparameters} summarizes the configuration for each dataset, including the number of transmitters $n_\text{tx}$, receivers $n_\text{rx}$, and the number of photons used for training. Photons are uniformly sampled from scene surfaces using the PyTorch3D library \cite{ravi2020pytorch3d}, ensuring broad angular coverage and surface diversity. Training is conducted using a batch size of $4 \times n_{\text{rx}}$. The radio frequency is fixed at 2.14 GHz across all experiments.}

\begin{table}[h]
\begin{center}
\caption{Hyperparameters of each dataset.}
\label{table: hyperparameters}
\begin{tabular}{|c|c|c|c|}
 \hline
 - & ``single building" & ``wi3rooms" & ``bistro" \\
 \hline
  photons & $2000$ & $3000$ & $9000$ \\
 \hline
 $n_\text{tx}$ (train) & $400$ & $400$ & $184$ \\
 \hline
 $n_\text{tx}$ (test) & $25$ & $25$ & $29$ \\
 \hline
 $n_\text{rx}$ & \makecell{$2700$ \\ $(30\times30\times3)$} & \makecell{$3750$ \\ $(50\times25\times3)$} & \makecell{$1000$ \\ (random)} \\
 \hline
 train epochs & $1500$ & $1000$ & $1000$ \\
 \hline
 learning rate & $0.0001$ & $0.0001$ & $0.0001$ \\
 \hline
\end{tabular}
\end{center}
\end{table}

\subsection{Single Building: Physical Fidelity and MIMO Application}

\ZVP{This experiment considers an outdoor scenario with a single building surrounded by an open space. It serves to validate the physical correctness, generalization capability, and real-time performance of Photon Splatting. Moreover, we discuss the application of the work in a multiuser MIMO setting.
}

\subsubsection{Environment Setup and Training pipeline}

\begin{figure}[!ht]
\centering
\subfigure[Training Configuration]{
    \begin{minipage}[t]{0.22\textwidth}
        \includegraphics[width=\linewidth]{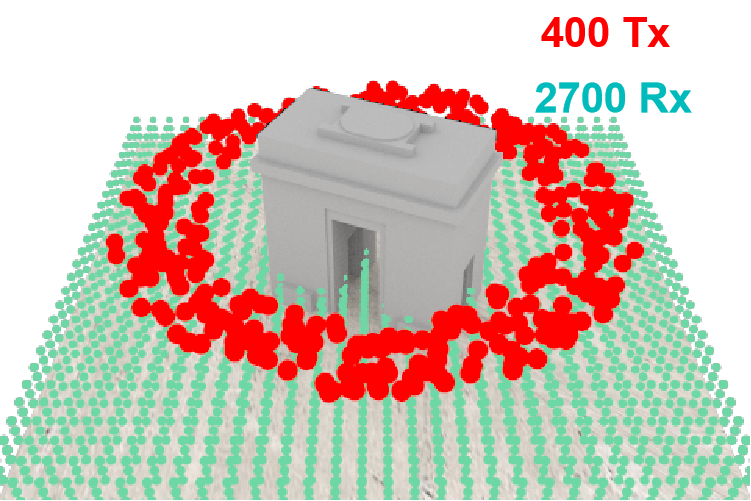}
        \label{subfig: single building Tx Rx plan (train)}
    \end{minipage}
}
\subfigure[Test Configuration]{
    \begin{minipage}[t]{0.22\textwidth}
        \includegraphics[width=\linewidth]{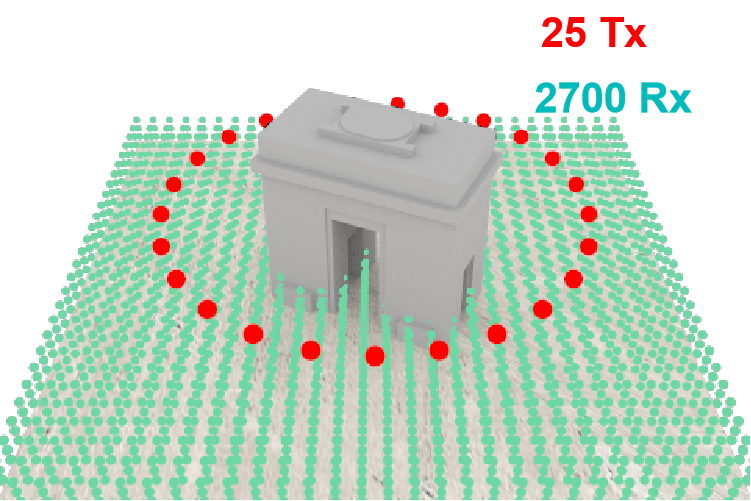}
        \label{subfig: single building Tx Rx plan (test)}
    \end{minipage}
}
\subfigure[Ray Tracing: Multipath Components]{
    \begin{minipage}[t]{0.4\textwidth}
        \includegraphics[width=\linewidth]{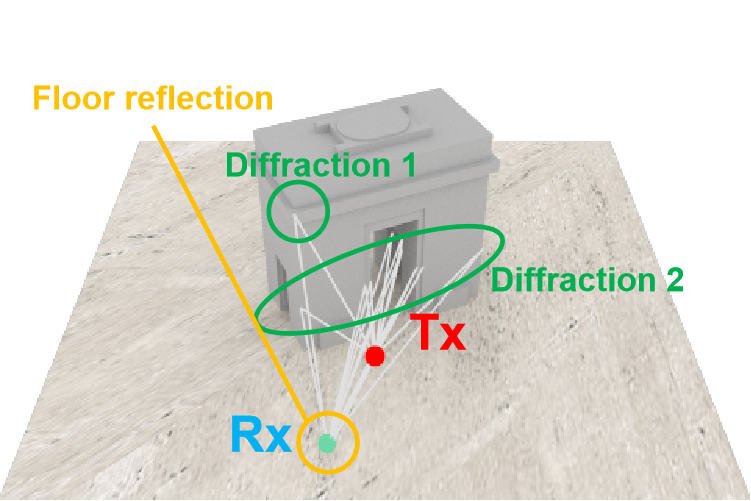}
        \label{subfig: ray tracing paths visualization}
    \end{minipage}
}
\subfigure[Photon Splatting: Directional Contributions]{
    \begin{minipage}[t]{0.48\textwidth}
        \includegraphics[width=\linewidth]{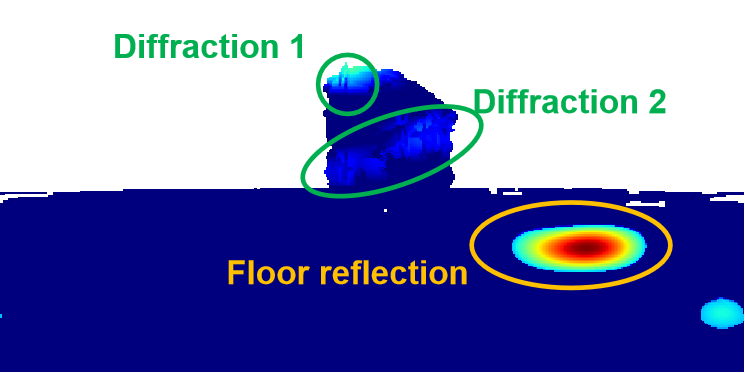}
        \label{subfig: photon visualization}
    \end{minipage}
}
\caption{
\ZVP{Single building dataset setup and photon visualization (Section IV.B.1).}  
}
\label{fig: single building photons}
\end{figure}

\ZVP{As illustrated in Fig.~\ref{fig: single building photons}, the scene consists of a single rectangular building placed on a flat outdoor surface.  It provides a clean yet physically rich setting to study wave propagation in outdoor environments, such as reflection off vertical walls, floor bounce, and edge diffraction. 

The training dataset is constructed to reflect practical scenarios in which UAVs provide aerial wireless coverage around physical structures.  Specifically, 400 transmitters (Tx) locations are uniformly sampled along a circular trajectory that encloses the building, as visualized in Fig.~\ref{subfig: single building Tx Rx plan (train)}. This configuration mimics UAV-mounted base stations orbiting the structure, providing coverage from a variety of angles and locations. The receiver grid comprises 2,700 uniformly placed points surrounding the building, emulating a dense distribution of mobile users on the ground. The test dataset is constructed by selecting a distinct set of transmitter positions, not seen during training (Fig.~\ref{subfig: single building Tx Rx plan (test)}). 
 
}

\ZVP{In Fig.~\ref{subfig: ray tracing paths visualization}, we plot a sample ray-tracing output between a Tx-Rx pair, highlighting the key physical interactions such as two diffraction events and a floor reflection. These rays represent the non-line-of-sight (NLOS) multipath components in this outdoor building scenario. In Fig.~\ref{subfig: photon visualization}, we show a visualization of the photon splatting prediction, rendered from the receiver's perspective using a geodesic spherical rasterizer. Each angular bin in the raster corresponds to a direction relative to the receiver, and the color intensity represents the inverse of the photon travel distance, i.e., shorter paths yield higher weights. Brighter regions in the plot indicate stronger directional contributions from surface-attached photons, which correspond well to the dominant multipath paths seen in the ray tracing output. This directional energy map illustrates how the model implicitly learns and captures physically meaningful multipath structures without explicitly tracing rays.}

\subsubsection{Validation on CIR Prediction and Spatial Coverage}

\begin{figure}[!ht]
\centering
\subfigure[CIR trajectory: 65 Rx locations captured along a pedestrian path around the building.]{
    \begin{minipage}[t]{0.48\textwidth}
        \includegraphics[width=\linewidth]{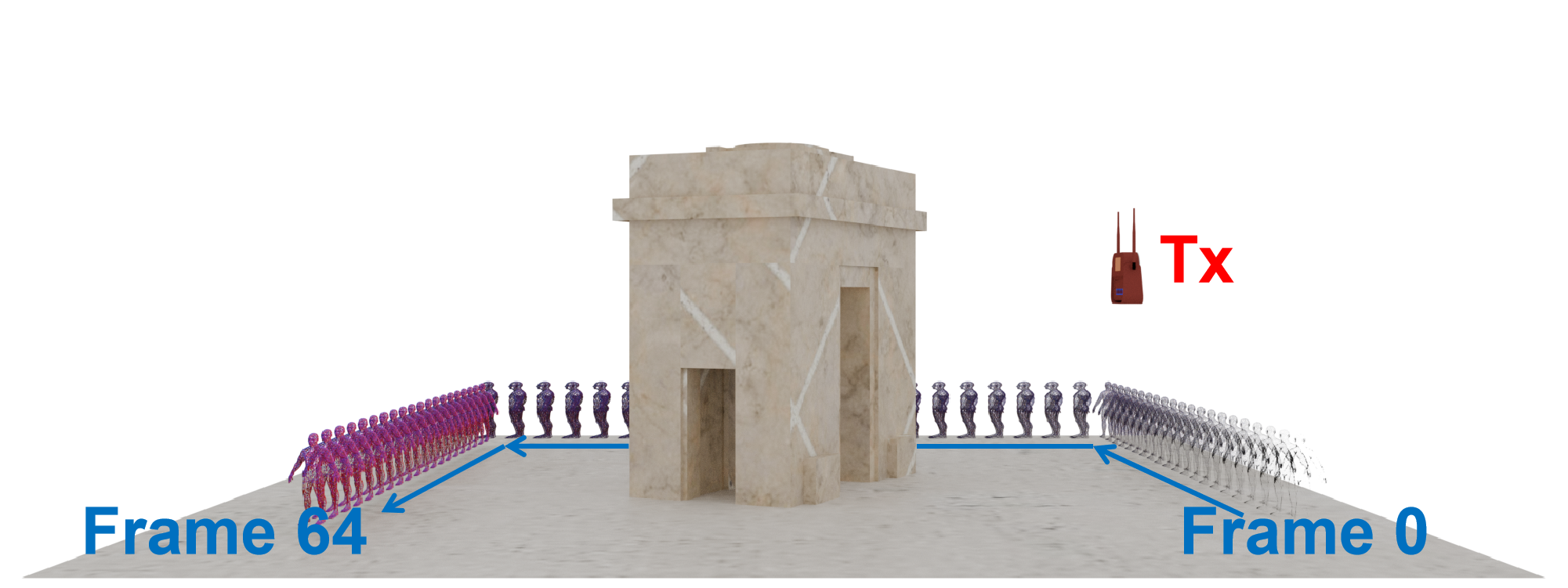}
        \label{subfig: single building CIR frames}
    \end{minipage}
}
\subfigure[Ground truth CIR at each Rx location.]{
    \begin{minipage}[t]{0.22\textwidth}
        \includegraphics[width=\linewidth]{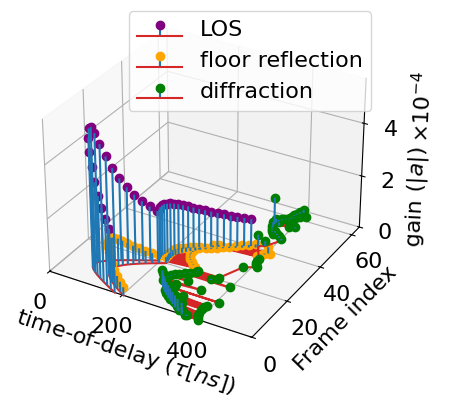}
        \label{subfig: single building CIR gt}
    \end{minipage}
}
\subfigure[Photon Splatting prediction at each Rx location.]{
    \begin{minipage}[t]{0.22\textwidth}
        \includegraphics[width=\linewidth]{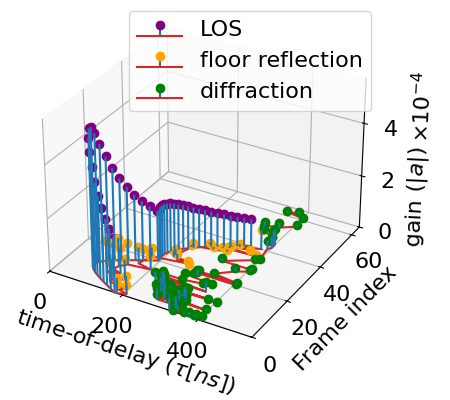}
        \label{subfig: single building CIR ours}
    \end{minipage}
}
\subfigure[Spatial coverage map over $30 \times 30 = 900$ Rx grid points.]{
    \begin{minipage}[t]{0.22\textwidth}
        \includegraphics[width=\linewidth]{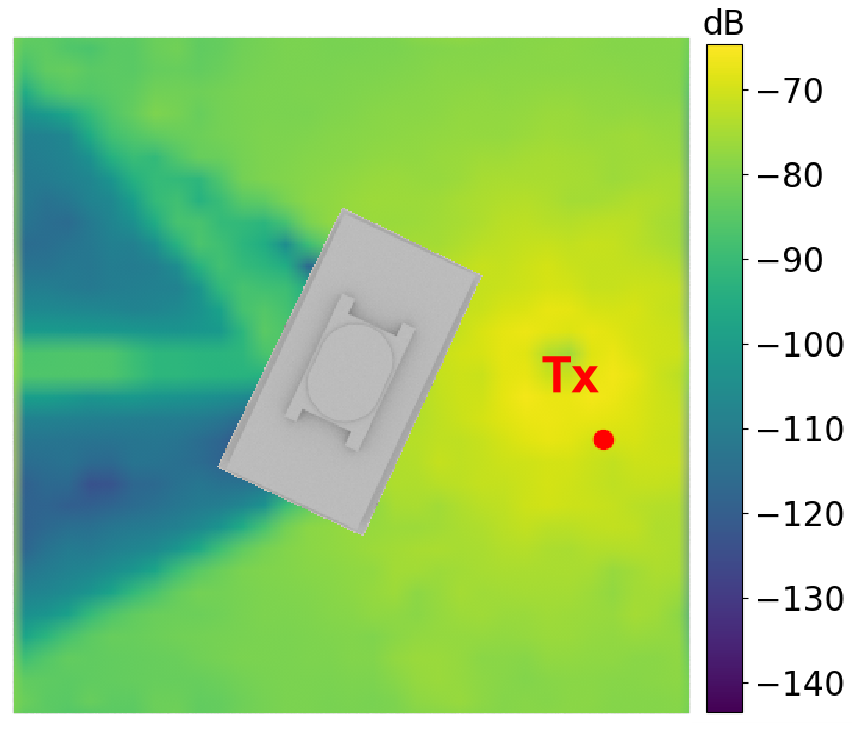}
        \vspace{0.5em}
        \parbox{\textwidth}{\centering \footnotesize Ground truth}
    \end{minipage}
    \begin{minipage}[t]{0.22\textwidth}
        \includegraphics[width=\linewidth]{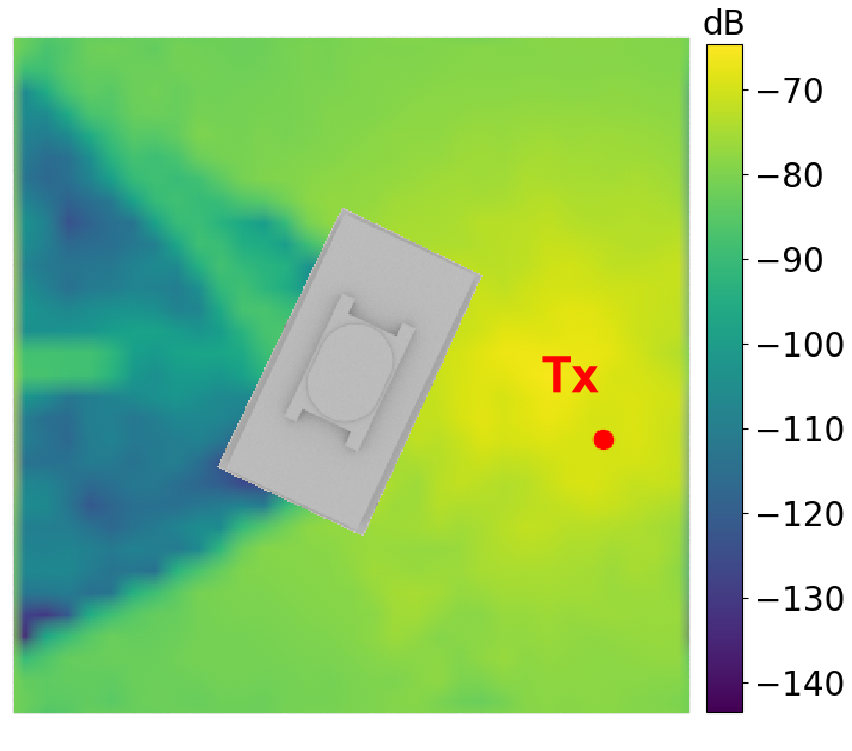}
        \vspace{0.5em}
        \parbox{\textwidth}{\centering \footnotesize Photon Splatting}
    \end{minipage}
    \label{subfig: single building cm}
}
\caption{Single building: CIR and spatial prediction (Section IV.B.2).}
\label{fig: single building}
\end{figure}

\ZVP{
To evaluate the prediction accuracy of our model, we select a transmitter configuration not seen during training and place multiple receivers around the building perimeter. This setup emulates a pedestrian trajectory that encircles the structure, enabling the study of time-varying wireless channel responses. At each receiver position, the model predicts the full  CIR, including the amplitude and delay of multipath components.

Figs~\ref{subfig: single building CIR frames} - \ref{subfig: single building CIR ours} compare the predicted CIRs with the ray-tracing-based ground truth across 65 successive receiver locations. In these 3D visualizations, the x-axis corresponds to path delay (time-of-arrival), the y-axis indexes the receiver positions along the trajectory, and the z-axis represents the amplitude of each multipath component. The comparison shows good agreement between prediction and ground truth, with both capturing the line-of-sight component, strong reflections from the floor, and weaker multipath arising from wall and edge interactions. This demonstrates the model's ability to reproduce complex, temporally varying channel behavior with high physical fidelity.

In addition to accuracy, we evaluate the model's performance under high-throughput conditions by predicting CIRs for 900 receivers arranged in a dense spatial grid. This dense prediction is completed in approximately 30 milliseconds, equivalent to 29 frames per second (FPS) on a single GPU, demonstrating the model's real-time capability for interactive wireless applications. The resulting spatial coverage map, visualized using the predicted channel frequency response $H(f)$, is shown in Fig.~\ref{subfig: single building cm}. The predicted map closely matches the ray-tracing ground truth, correctly identifying regions of signal enhancement and shadowing caused by building reflections and obstructions.  These results confirm that Photon Splatting enables fast and physically consistent prediction of spatial coverage maps across dense receiver grids, supporting real-time applications such as adaptive beamforming, localization-aware signal planning, and channel-aware control in wireless digital twin systems.
}

\subsubsection{Multiuser MIMO Application via Channel Precoding}

\begin{figure}[!ht]
\centering
\subfigure[GT: $\mathbf{H}^{H} \mathbf{W} = \mathbf{I}$]{
    \begin{minipage}[t]{0.22\textwidth}
        \includegraphics[width=\linewidth]{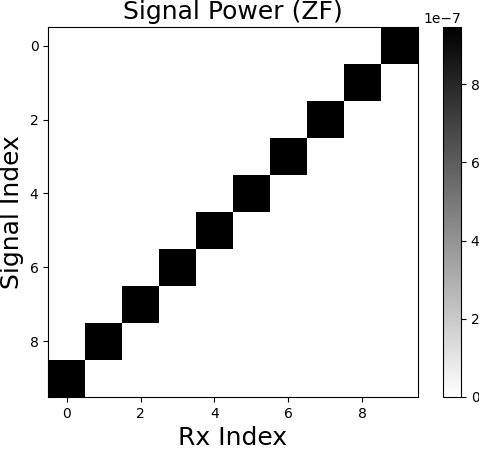}
        \label{subfig: MIMO precoding GT}
    \end{minipage}
}
\subfigure[Ours: $\mathbf{H}^{H} \mathbf{W} \sim \mathbf{I}$]{
    \begin{minipage}[t]{0.22\textwidth}
        \includegraphics[width=\linewidth]{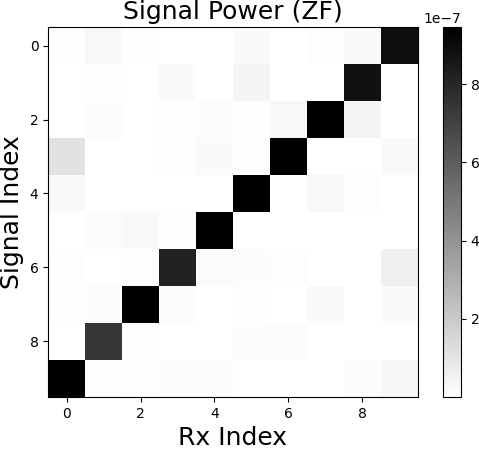}
        \label{subfig: MIMO precoding Ours}
    \end{minipage}
}
\caption{Multiuser MIMO Precoding Validation (Section IV.B.3)
}
\label{fig: single building MIMO precoding}
\end{figure}

\ZVP{
To evaluate the practical utility of Photon Splatting (PS) in wireless systems, we simulate a multiuser MIMO downlink scenario with 25 transmitters and 10 receivers. This setup emulates a UAV swarm base station serving mobile users around a building.

We use the predicted channel frequency response matrix \( \mathbf{H}_{\text{PS}}(f) \in \mathbb{C}^{n_\text{rx} \times n_\text{tx}} \) to compute the zero-forcing precoding matrix:
\[
\mathbf{W}_{\text{PS}} = \mathbf{H}_{\text{PS}}^{H} \left( \mathbf{H}_{\text{PS}} \mathbf{H}_{\text{PS}}^{H} \right)^{-1}.
\]
The resulting downlink transmission is modeled as:
\[
\mathbf{y} = \mathbf{H}_{\text{GT}} \mathbf{W}_{\text{PS}} \mathbf{x} + \mathbf{n},
\]
where \( \mathbf{H}_{\text{GT}} \) is the ray-tracing-based ground-truth channel used for evaluation only, \( \mathbf{x} \) is the transmit symbol vector, and \( \vec{n} \) is additive noise.

For reference, we also compute an ideal precoding matrix using the ground-truth:
\[
\mathbf{W}_{\text{GT}} = \mathbf{H}_{\text{GT}}^{H} \left( \mathbf{H}_{\text{GT}} \mathbf{H}_{\text{GT}}^{H} \right)^{-1}.
\]
In both cases, effective precoding is indicated when the matrix product \( \mathbf{H}_{\text{GT}} \mathbf{W} \) approximates the identity matrix.

As shown in Fig.~\ref{fig: single building MIMO precoding}, the PS-based precoder yields a result that closely resembles that of the ground-truth, validating that the predicted channels are sufficiently accurate for real-world physical-layer tasks such as spatial multiplexing and interference suppression.

More significantly, this experiment supports a new operational paradigm: when the scene geometry and user/device locations are known, Photon Splatting enables direct computation of precoding matrices \emph{without channel estimation}. 
This eliminates the need for costly pilot transmission and estimation overhead, offering a compelling solution for latency-sensitive and bandwidth-constrained applications such as UAV swarms, robotics, and mmWave communication.
}

\subsection{Indoor Experiment: Scalability, Generalization, and Neural Architecture Evaluation}

\begin{figure*}[!t]
\centering
\subfigure[Bistro (physical): A cafe]{
    \begin{minipage}[t]{0.22\textwidth}
      \includegraphics[width=\linewidth]{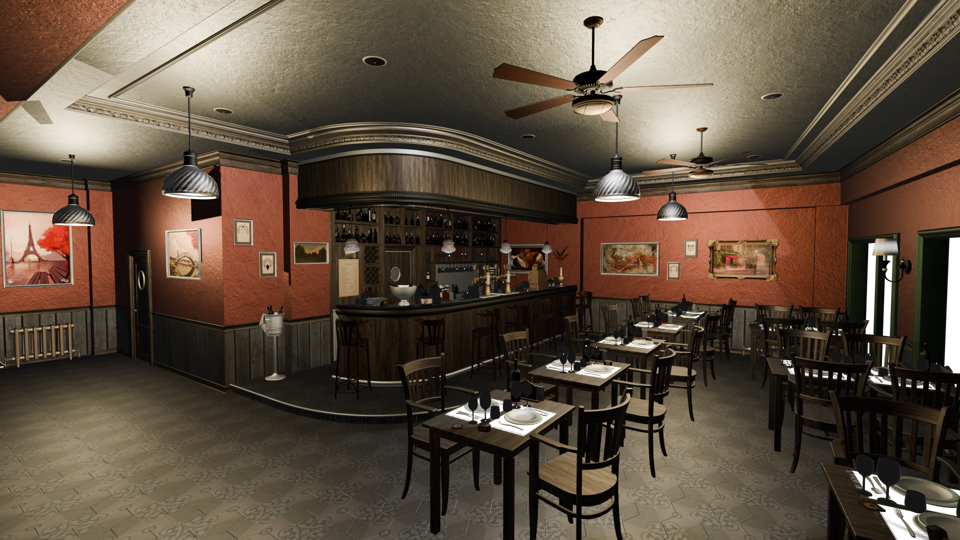}
        \label{fig: bistro view}
    \end{minipage}
}
\subfigure[Bistro (virtual): photons]{
    \begin{minipage}[t]{0.22\textwidth}
     \includegraphics[width=\linewidth]{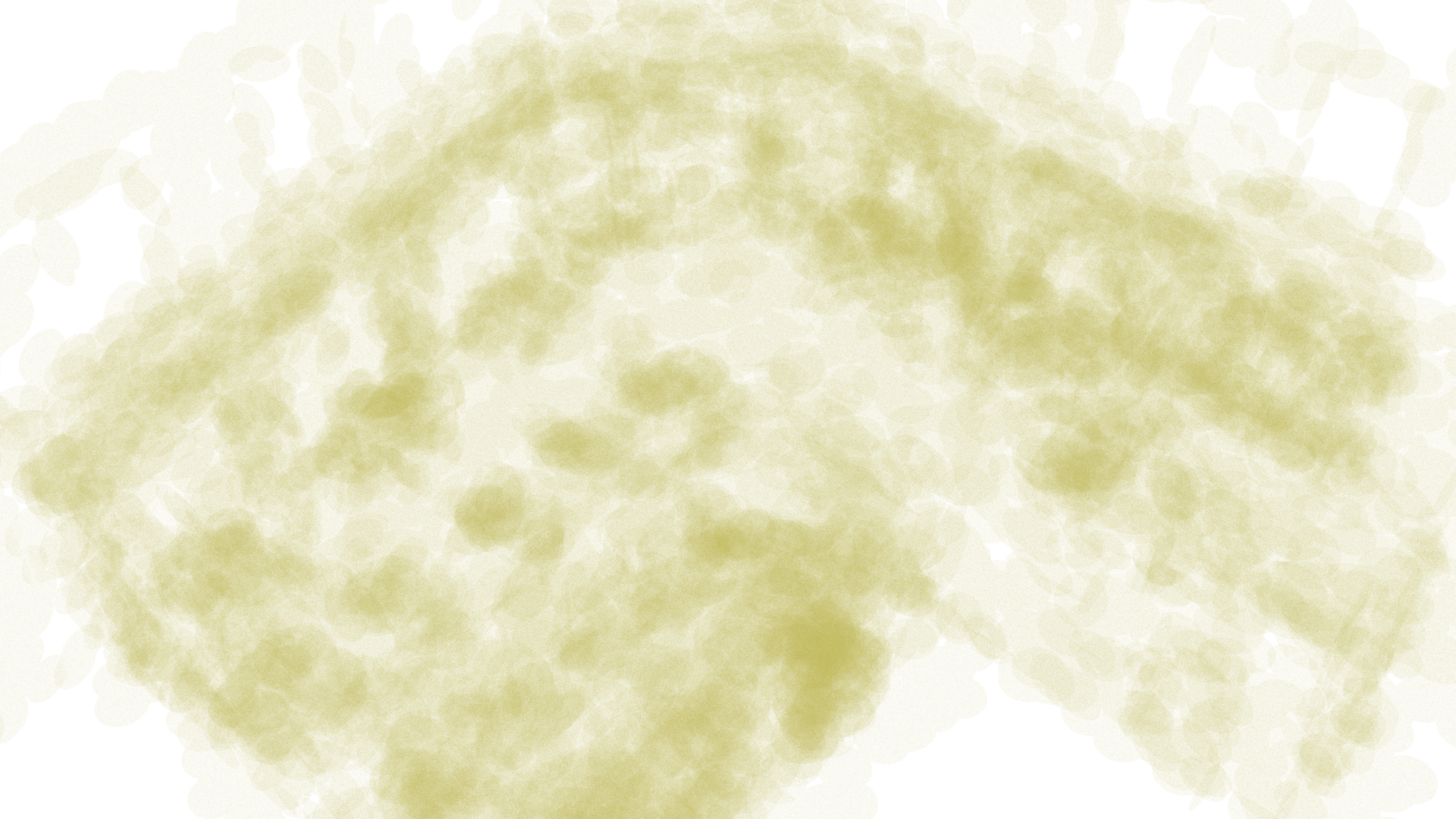}
        \label{fig: bistro photons}
    \end{minipage}
}
\subfigure[Ground Truth (Tx $\# 15$)]{
    \begin{minipage}[t]{0.22\textwidth}
      \includegraphics[width=\linewidth]{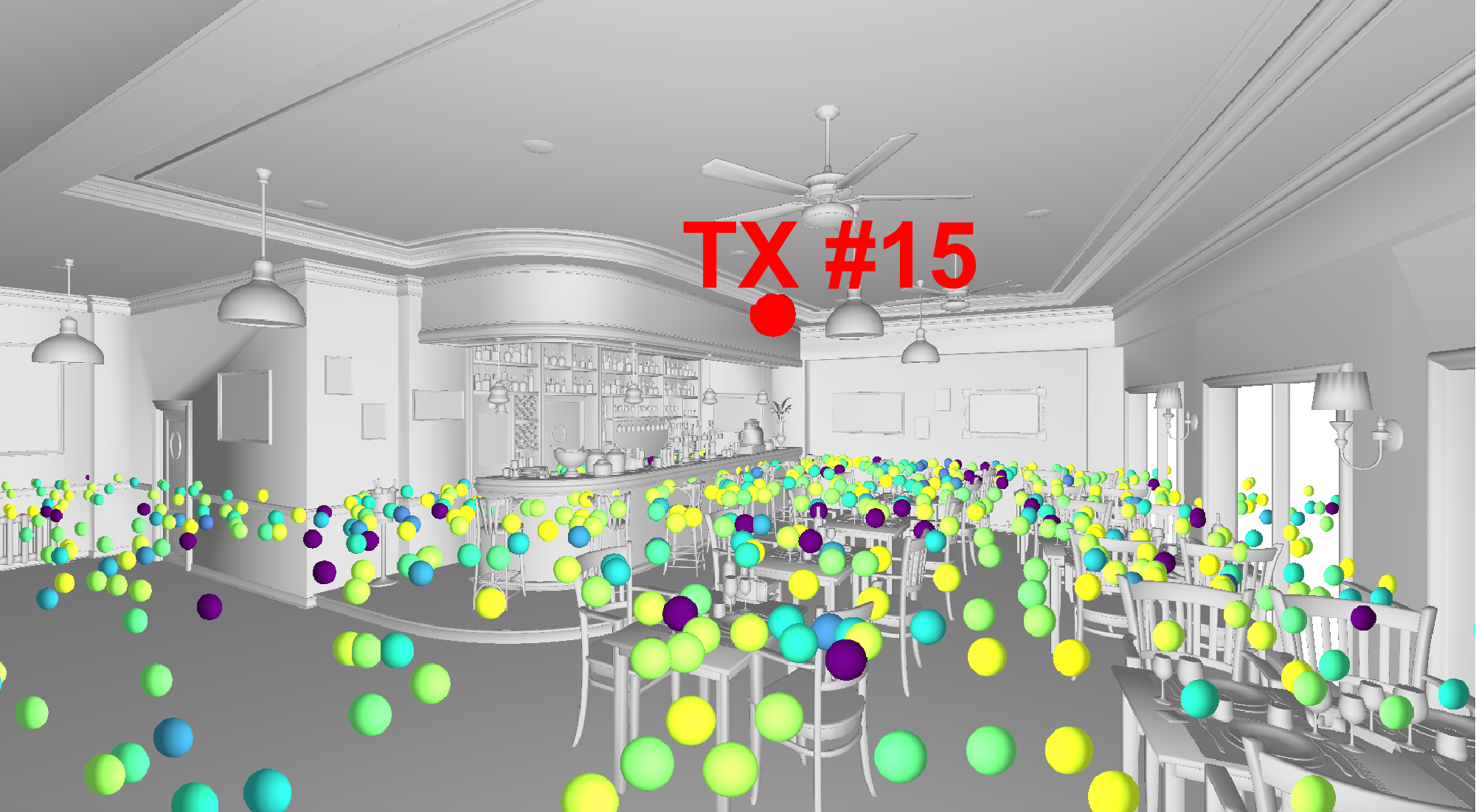}
        \label{fig: bistro gui gt}
    \end{minipage}
}
\subfigure[Photon Splatting (Tx $\# 15$)]{
    \begin{minipage}[t]{0.22\textwidth}
      \includegraphics[width=\linewidth]{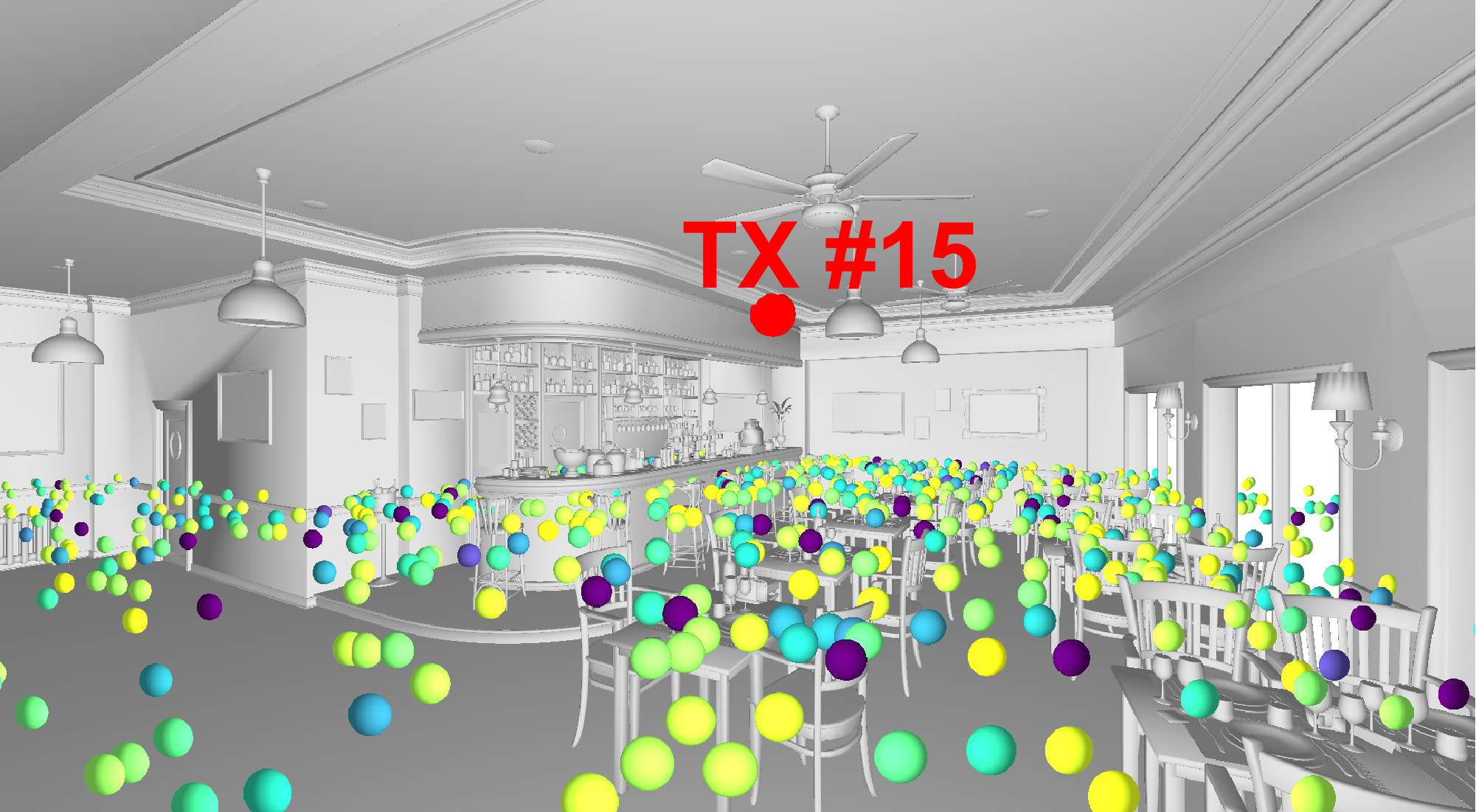}
        \label{fig: bistro gui pred}
    \end{minipage}
}
\subfigure[Tx/Rx planning (test)]{
    \begin{minipage}[t]{0.22\textwidth}
      \includegraphics[width=\linewidth]{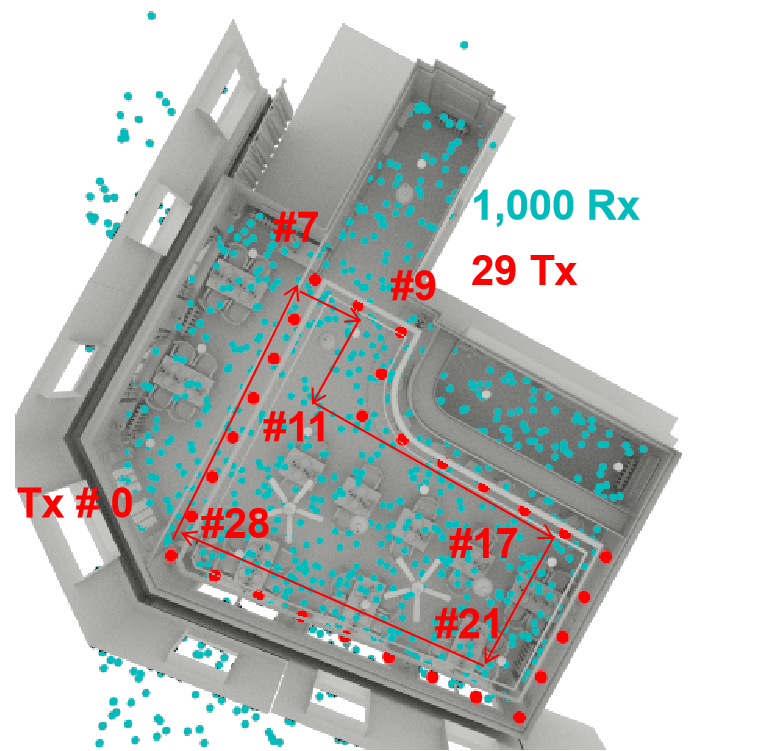}
        \label{fig: bistro devices}
    \end{minipage}
}
\subfigure[Average channel gain]{
    \begin{minipage}[t]{0.22\textwidth}
      \includegraphics[width=\linewidth]{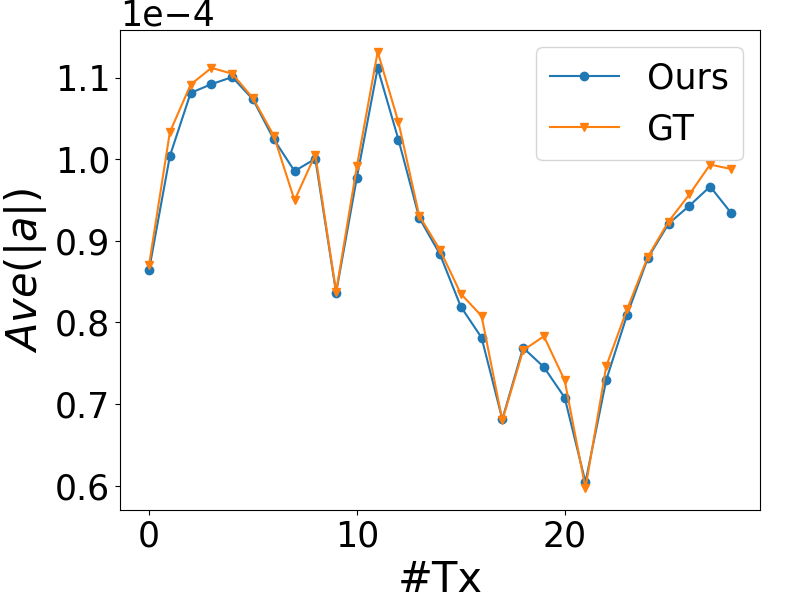}
        \label{fig: bistro optim}
    \end{minipage}
}
\subfigure[Ground Truth (best Tx)]{
    \begin{minipage}[t]{0.22\textwidth}
      \includegraphics[width=\linewidth]{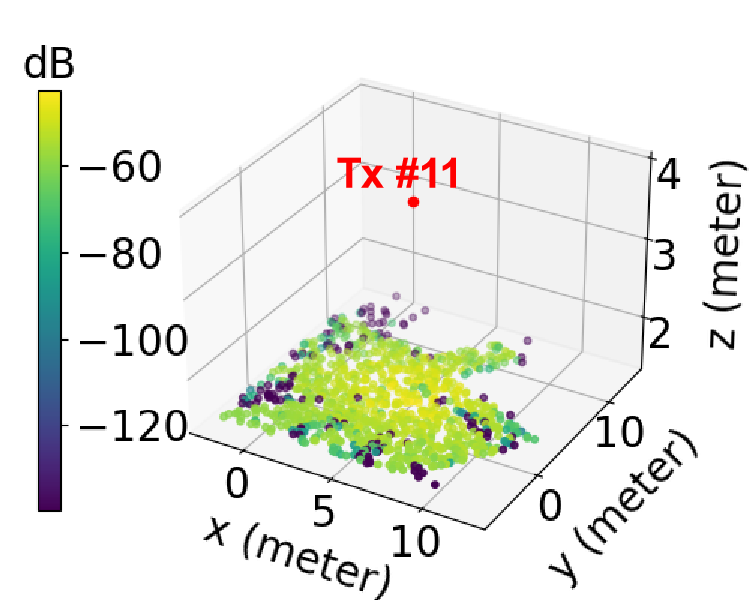}
        \label{fig: bistro gt}
    \end{minipage}
}
\subfigure[Photo Splatting (best Tx)]{
    \begin{minipage}[t]{0.22\textwidth}
      \includegraphics[width=\linewidth]{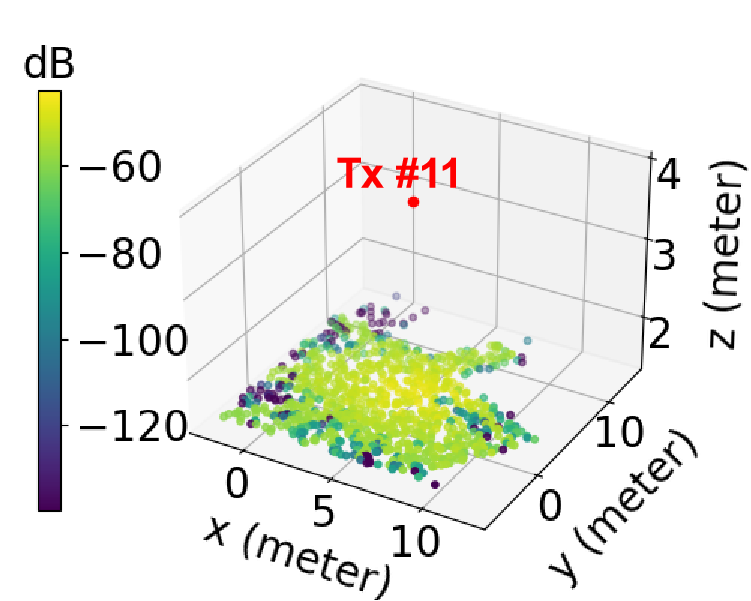}
        \label{fig: bistro pred}
    \end{minipage}
}
\caption{Bistro Scene and Prediction Overview. 
\subref{fig: bistro view} - \subref{fig: bistro photons} Visualization of the physical Bistro scene and surface-attached photons.
\subref{fig: bistro devices} The placement of 1,000 randomly distributed Rxs and 29 ceiling-mounted test Txs. The training Tx locations are not shown due to image constraints. \subref{fig: bistro gui gt}/\subref{fig: bistro gui pred} Rx's signal strength for a selected test Tx, the colorful spheres denotes the positions and received signal strength for each Rx. \subref{fig: bistro gt}/\subref{fig: bistro pred} Photon Splatting prediction for the same Tx, showing close agreement with the ground truth. \subref{fig: bistro optim} The comparison of the average channel gain across 1000 Rx for the 29 testing Tx locations.
}
\label{fig: bistro}
\end{figure*}

\begin{figure*}
    \centering
    \renewcommand{\arraystretch}{1.5} 
    \begin{tabular}{cc|cc}  
        \multicolumn{2}{c|}{\underline{Pattern ``Tr389.01" was \textbf{exposed} on training process.}} &  
        \multicolumn{2}{c}{\underline{Pattern ``Half-wavelength diople" is \textbf{unseen} from training.}} \\
         & & & \\
        
        \includegraphics[width=0.22\textwidth]{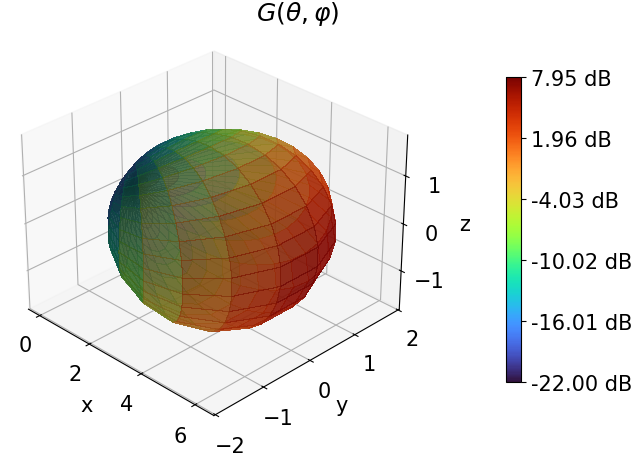} & 
        \includegraphics[width=0.22\textwidth]{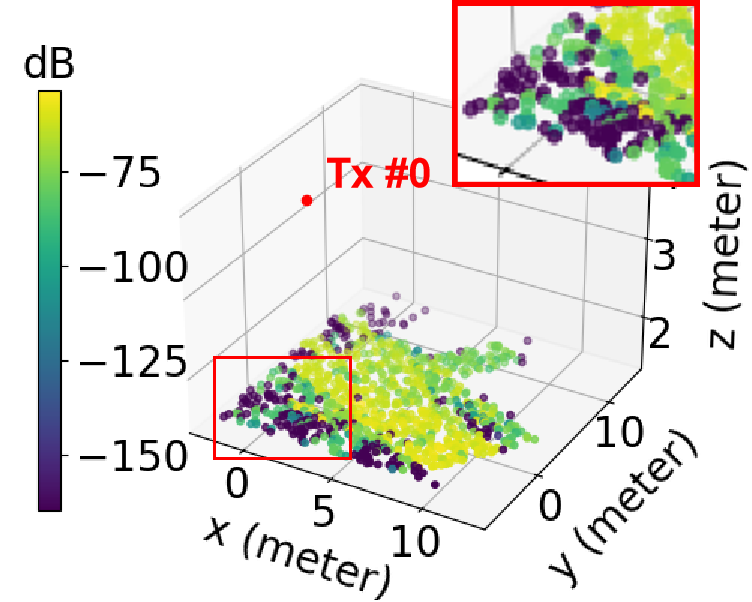} & 
        \includegraphics[width=0.22\textwidth]{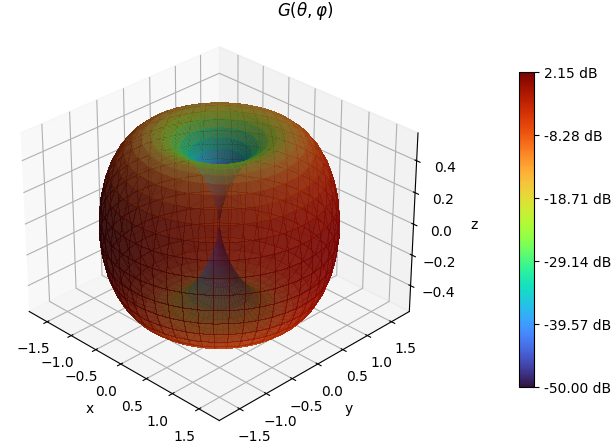} & 
        \includegraphics[width=0.22\textwidth]{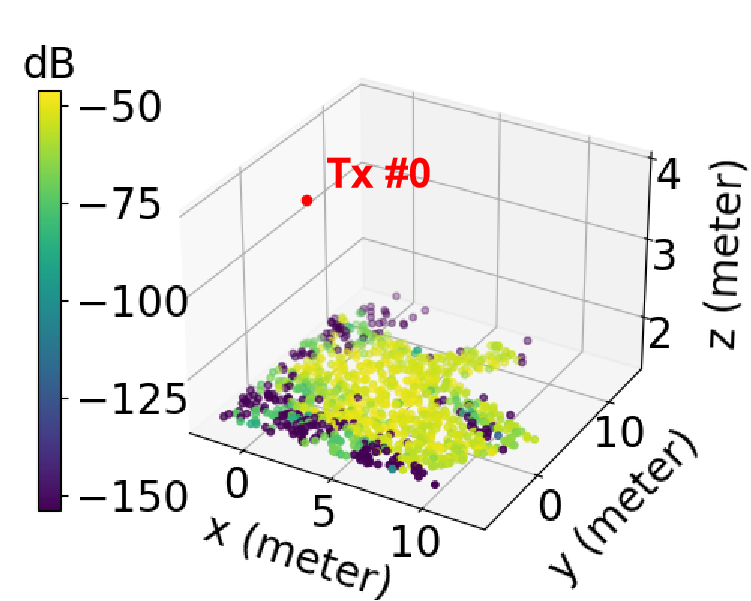} \\
        (a) Antenna pattern: Tr389.01 & (b) Ray-tracing (\textbf{GT}) & (g) Pattern: HW dipole & (h) Ray-tracing (\textbf{GT}) \\

        \includegraphics[width=0.22\textwidth]{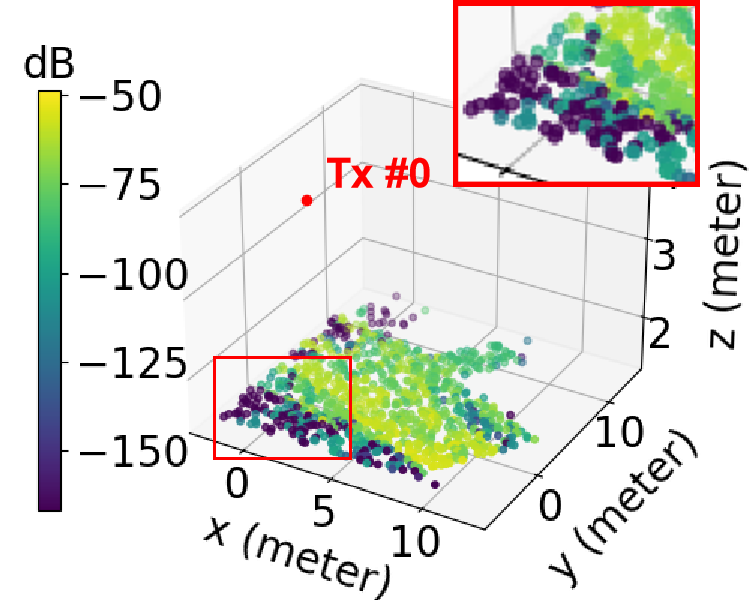} & 
        \includegraphics[width=0.22\textwidth]{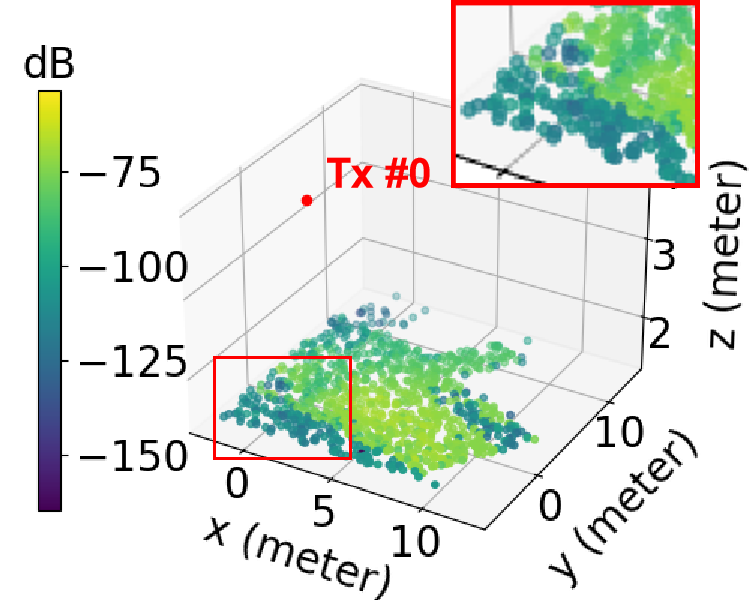} & 
        \includegraphics[width=0.22\textwidth]{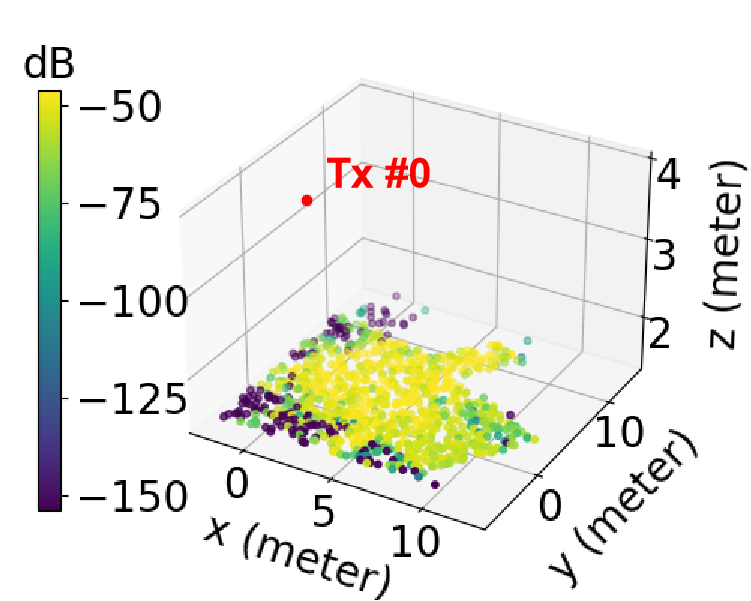} & 
        \includegraphics[width=0.22\textwidth]{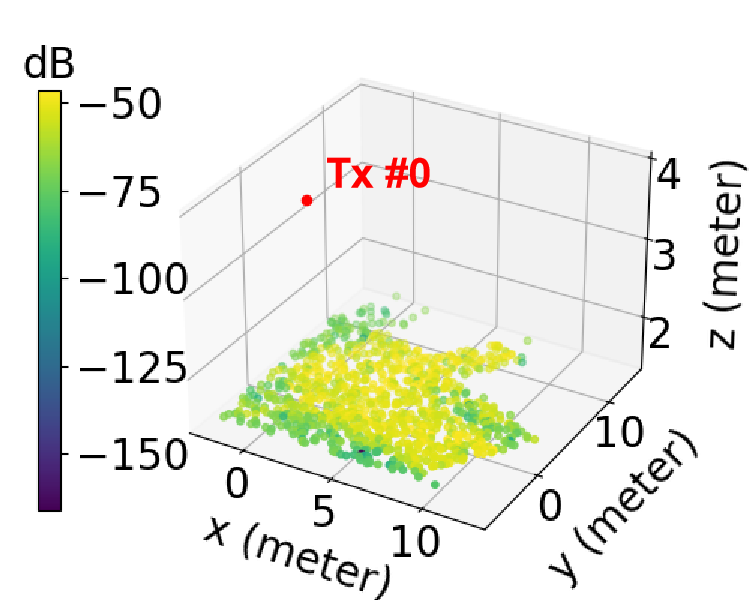} \\
        (c) Photon-Splatting w/ FNO & (d) Photon-Splatting w/ & (i) Photon-Splatting w/ FNO & (j) Photon-Splatting w/ \\
        (\textbf{Ours}) & Attention (\textbf{Ablation}) & (\textbf{Ours}) & Attention (\textbf{Ablation}) \\

        \includegraphics[width=0.22\textwidth]{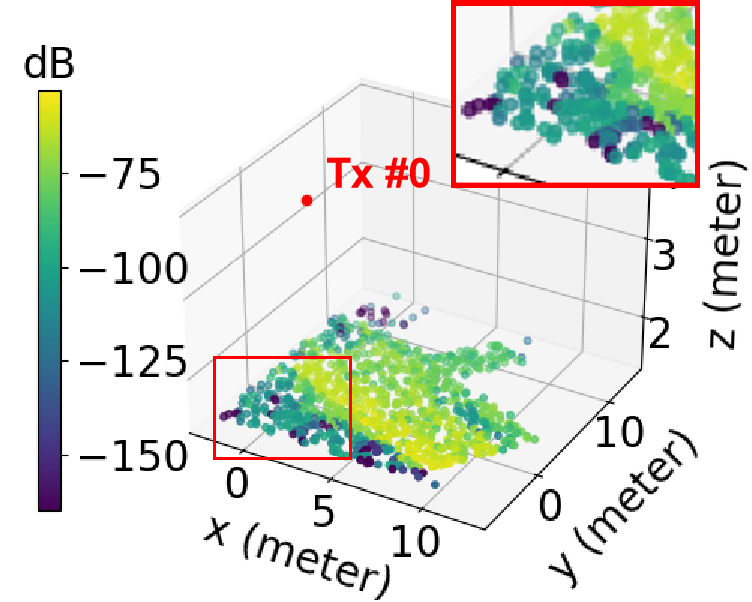} & 
        \includegraphics[width=0.22\textwidth]{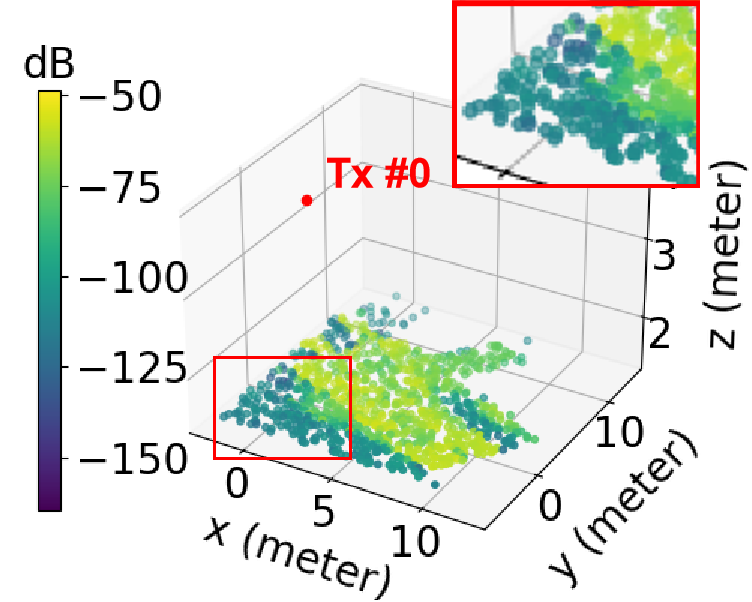} & 
        \includegraphics[width=0.22\textwidth]{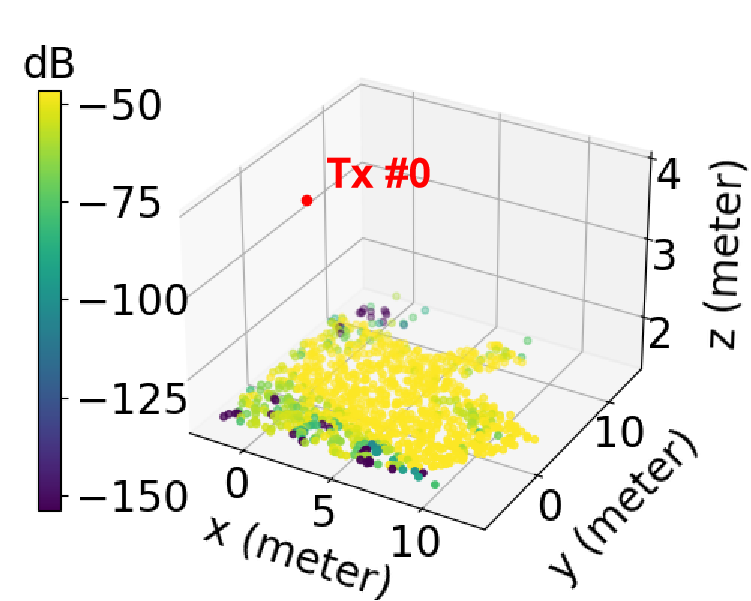} & 
        \includegraphics[width=0.22\textwidth]{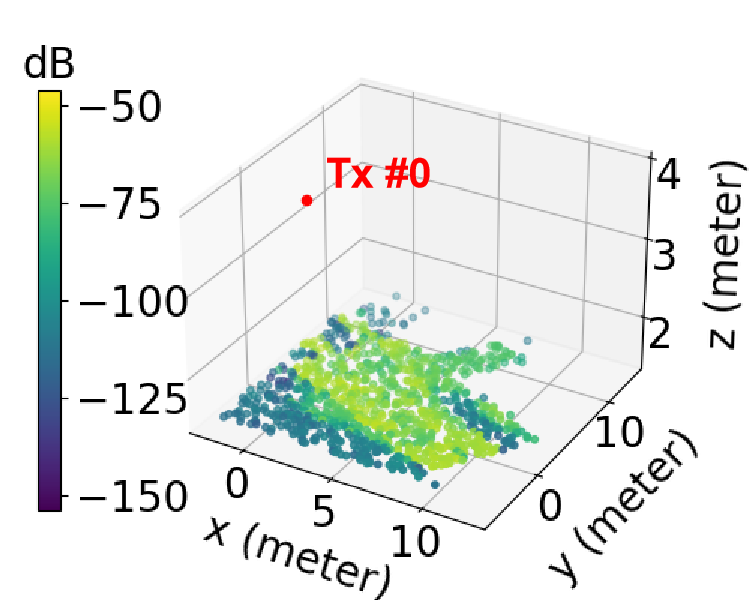} \\
        (e)  Photon-Splatting w/ MLP & (f) MLP & (k)  Photon-Splatting w/ MLP & (l) MLP \\
        (\textbf{Ablation}) & (\textbf{Comparison}) & (\textbf{Ablation}) & (\textbf{Comparison}) \\
    \end{tabular}

    \caption{{Ablation Experiments and Comparison (Section IV.C.3)}. Using the ``bistro" caf\'e dataset, we conduct ablation studies by replacing our original neural encoder, ``FNO" (c), with ``Attention" (c) and ``MLP" (d). Additionally, we compare our approach against a pure MLP pipeline (f). The ground truth, obtained from a ray-tracing algorithm, is presented in (b). All of these models are trained solely under ``Tr389.01" antenna pattern (a)-(f). Additionally, we also provide the test on another unseen pattern ``Half-wavelength dipole" (g)-(l). The results show that Photon Splatting with FNO better captures spatial signal patterns and generalizes more effectively to new antenna types.}
    \label{fig: bistro ablation}
\end{figure*}

\begin{table*}
\begin{center}
\caption{Ablation and Comparison Results on the Bistro Dataset. We evaluate the performance of Photon Splatting variants using three metrics: relative MSE for path loss, average delay error AE($\tau$), and PSNR. The upward arrow indicates better performance with larger values, while the downward arrow denotes better performance with smaller values. Bold entries denote the best-performing models across each metric.}
\label{table: ablation}
\begin{tabular}{|c|c|c|c|c|}
 \hline
 models & relative MSE $\downarrow$ & AE ($\tau$) [ns] $\downarrow$ & PSNR $\uparrow$ & \#train parameters \\
 \hline
 Photon Splatting w/ \textbf{FNO} (\textbf{Proposed}) & $\boldsymbol{0.023}$ & $\boldsymbol{1.95}$ & $\boldsymbol{16.58}$ & $2,427,387$\\
  \hline
 Photon Splatting w/ Attention (\textbf{Ablation}) & $0.033$ & $4.53$ & $15.07$ & $3,587,656$\\
 \hline
 Photon Splatting w/ MLP (\textbf{Ablation}) & $0.024$ & $3.48$ & $16.39$ & $2,039,163$\\
 \hline
 MLP (\textbf{Comparison}) & $0.026$ & $4.18$ & $16.20$ & $1,787,285$\\
 \hline
\end{tabular}
\end{center}
\end{table*}

\ZVP{
This experiment evaluates the performance of Photon Splatting in a complex, realistic indoor environment using the publicly available ``bistro'' scene \cite{ORCAAmazonBistro}. The goal is to validate three key capabilities: (i) physical fidelity in densely occluded environments, (ii) practical utility in large-scale transmitter placement and selection, and (iii)  generalization to previously unseen antenna beam patterns.
}

\ZVP{
\subsubsection{Scene Description and Dataset Configuration}
The Bistro scene captures a furnished caf\'e interior with detailed furnishing, including walls, windows, tables, chairs, and other architectural elements (Fig.~\ref{fig: bistro view}). This makes it a challenging benchmark for high-fidelity channel modeling, as it introduces numerous occlusions, multi-bounce reflections, and complex shadow zones.

The training data is generated using the Sionna ray-tracing engine, where 9,000 photons are attached to interior surfaces (Fig.~\ref{fig: bistro photons}). The training dataset includes 184 randomly distributed Txs, while the test set consists of 29 new Txs mounted along the ceiling, reflecting practical indoor access point (AP) placements.  For each transmitter, 1,000 receivers (Rxs) are randomly scattered across the floor to emulate a dense user population. This setup enables rigorous evaluation of the model's ability to generalize to unseen transmitter configurations and accurately capture complex propagation dynamics in densely occluded indoor environments.
}

\ZVP{
\subsubsection{Prediction Quality and Runtime}

To evaluate the accuracy of Photon Splatting, we first visualize the predicted received signal strength over 1,000 randomly distributed receivers for a selected test transmitter. Figs.~\ref{fig: bistro gui gt} and~\ref{fig: bistro gui pred} compare the predicted spatial channel response to the ray-tracing ground truth. The color intensity of each sphere indicates the received signal amplitude at a receiver. The spatial distribution and magnitude predicted by our model closely match the ground truth result.

Next, we evaluate transmitter performance across the full test set by computing the average received signal amplitude over all 1,000 receivers for each of the 29 ceiling-mounted transmitters. As shown in Fig.~\ref{fig: bistro optim}, the model correctly identifies transmitter \#11 as providing the strongest coverage, while transmitter \#21 yields among the lowest. This emulates a real-world planning scenario in which system designers select optimal antenna placements for indoor coverage.
Finally, we show the strongest coverage (i.e. \#11) and its comparison with ground truth in Fig.~\ref{fig: bistro gt}-\subref{fig: bistro pred}.

We remark that the entire evaluation across 29 transmitters completes in under 1 second (0.98 seconds), corresponding to 0.03 seconds per transmitter configuration. In contrast, ray tracing requires over 147 seconds per transmitter configuration. This real-time throughput simulation enables practical use in wireless digital twin systems for antenna planning, coverage optimization, and interactive environment exploration.
}

\ZVP{
\subsubsection{Neural Architecture Evaluation}

To assess the impact of neural encoder design on predictive performance, we conduct a comparative evaluation of three Photon Splatting variants, each using a different encoder architecture: Fourier Neural Operator (FNO), Multi-Layer Perceptron (MLP), and an Attention-based module. All models are trained and evaluated under identical conditions, allowing for a controlled assessment of architectural differences.

Fig.~\ref{fig: bistro ablation} shows qualitative results for both a seen antenna pattern (TR389.01) and an unseen pattern (half-wavelength dipole). Among the three variants, the FNO-based model consistently produces more accurate spatial field distributions, particularly in capturing fine-scale multipath structures and angular variations. In contrast, MLP and Attention-based models tend to oversmooth or miss subtle spatial variations, especially when generalizing to unseen antenna configurations.

These observations are quantitatively supported by the results in Table~\ref{table: ablation}. Across three key metrics - relative mean squared error (MSE), average delay error AE($\tau$), and peak signal-to-noise ratio (PSNR) - the FNO variant achieves the best overall performance, with the lowest MSE (0.023), the most accurate delay prediction (1.95 ns), and the highest PSNR (16.58). This highlights the FNO's superior performance in capturing small-scale spatial correlations and wave-structure interactions in complex environments.
}

\subsection{Indoor Wi3rooms: Directional Delay Prediction and Robotics Trajectory Planning}

\begin{figure}[!t]
\centering
\subfigure[Room geometry]{
    \begin{minipage}[t]{0.22\textwidth}
     \includegraphics[width=\linewidth]{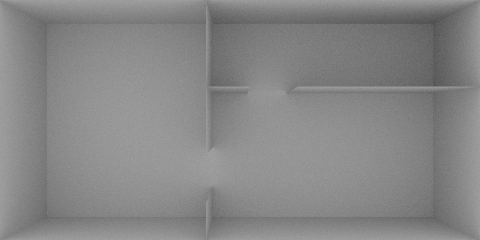}
        \label{subfig: room geometry}
    \end{minipage}
}
\subfigure[Tx/Rx planning (train dataset)]{
    \begin{minipage}[t]{0.22\textwidth}
     \includegraphics[width=\linewidth]{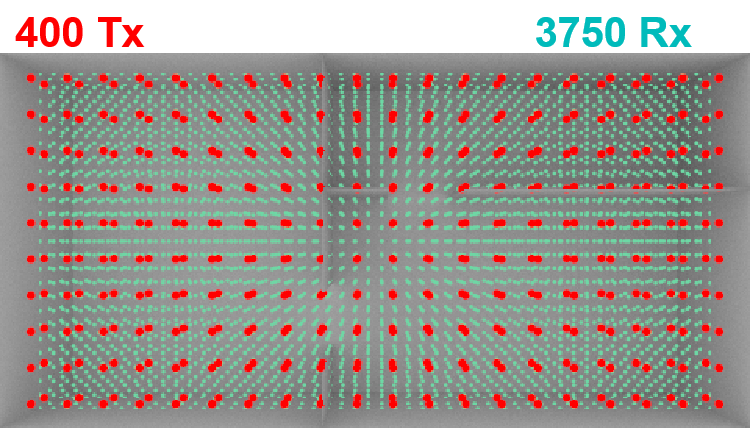}
        \label{subfig: wi3rooms Tx/Rx planning (train dataset)}
    \end{minipage}
}
\caption{\textbf{``wi3rooms" dataset (IV.D)}: We train our model under one antenna pattern, under the Tx/Rx settings of \subref{subfig: wi3rooms Tx/Rx planning (train dataset)}.}
\label{fig: wi3rooms}
\end{figure}

\begin{figure}[!t]
\centering
\subfigure[Ours: frame 0 (start position)]{
    \begin{minipage}[t]{0.22\textwidth}
      \includegraphics[width=\linewidth]{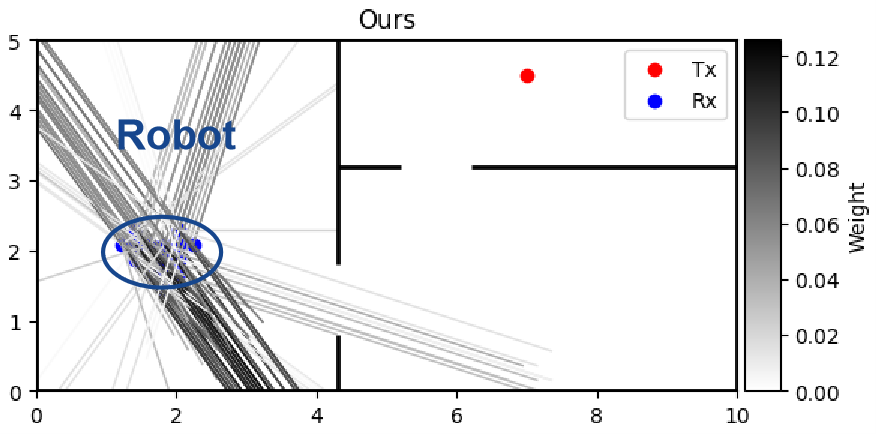}
        \label{fig: robotics frame 0 Ours}
    \end{minipage}
}
\subfigure[GT: frame 0 (start position)]{
    \begin{minipage}[t]{0.22\textwidth}
     \includegraphics[width=\linewidth]{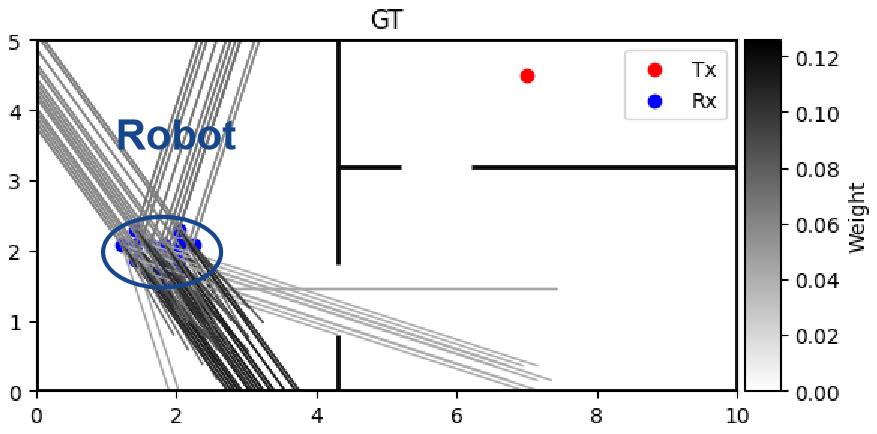}
        \label{fig: robotics frame 0 GT}
    \end{minipage}
}
\subfigure[Ours: frame 1 (found the door)]{
    \begin{minipage}[t]{0.22\textwidth}
      \includegraphics[width=\linewidth]{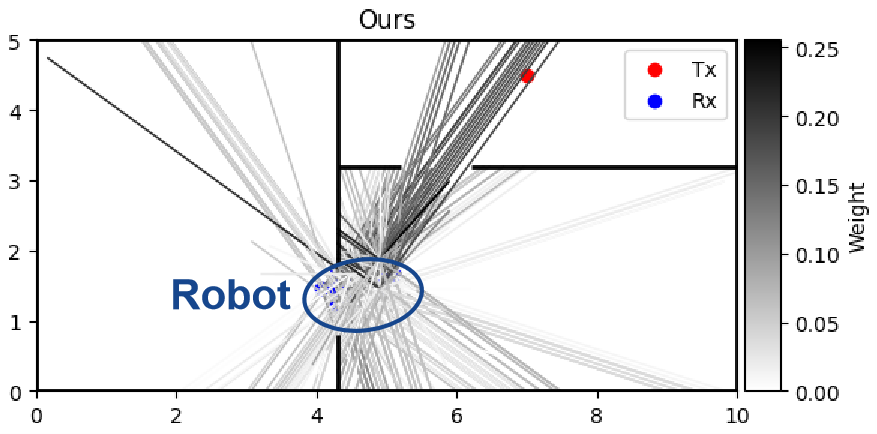}
        \label{fig: robotics frame 1 Ours}
    \end{minipage}
}
\subfigure[GT: frame 1 (found the door)]{
    \begin{minipage}[t]{0.22\textwidth}
     \includegraphics[width=\linewidth]{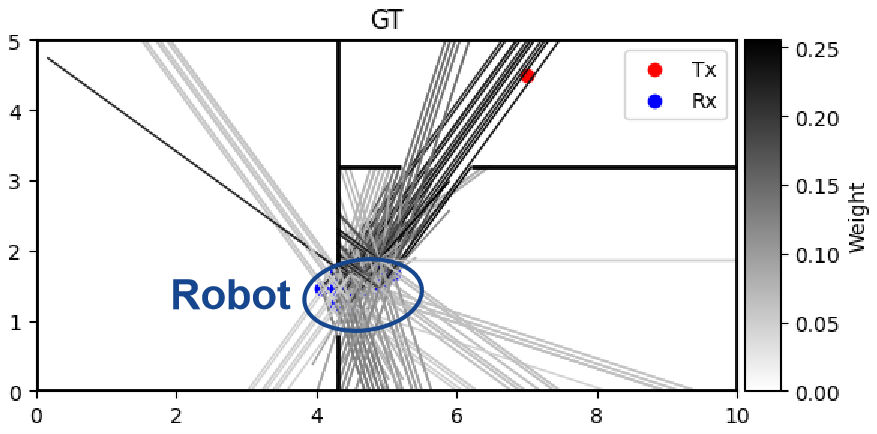}
        \label{fig: robotics frame 1 GT}
    \end{minipage}
}
\subfigure[Ours: frame 2 (target position)]{
    \begin{minipage}[t]{0.22\textwidth}
      \includegraphics[width=\linewidth]{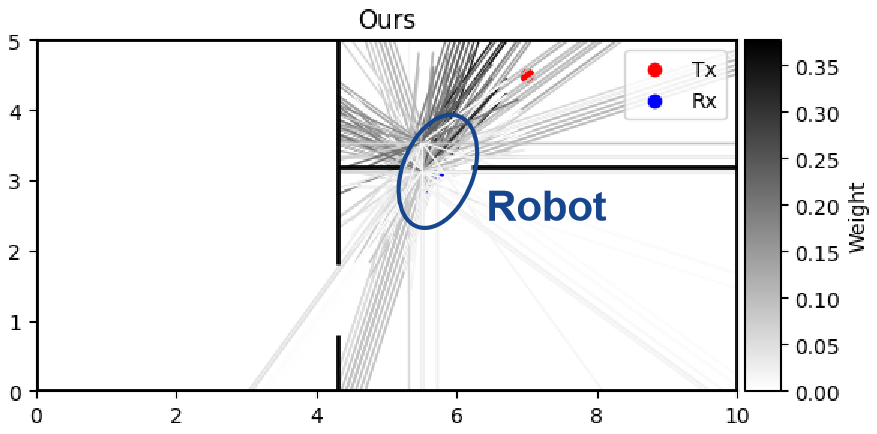}
        \label{fig: robotics frame 2 Ours}
    \end{minipage}
}
\subfigure[GT: frame 2 (target position)]{
    \begin{minipage}[t]{0.22\textwidth}
     \includegraphics[width=\linewidth]{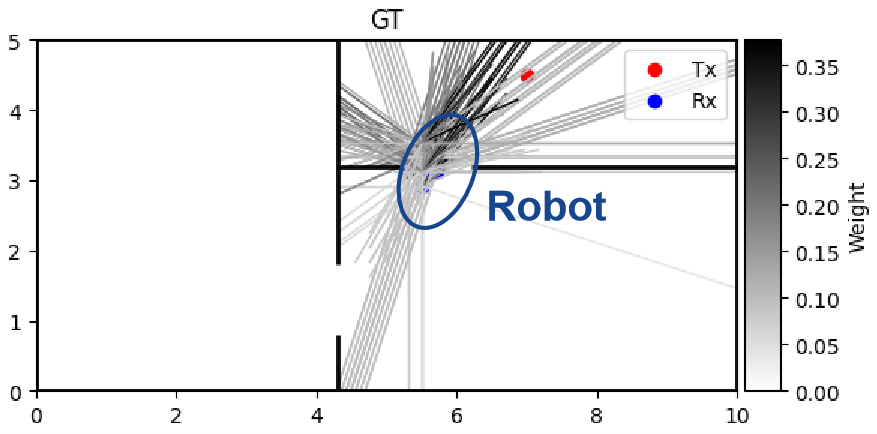}
        \label{fig: robotics frame 2 GT}
    \end{minipage}
}
\subfigure[Rendered frames (full trace)]{
    \begin{minipage}[t]{0.22\textwidth}
     \includegraphics[width=\linewidth]{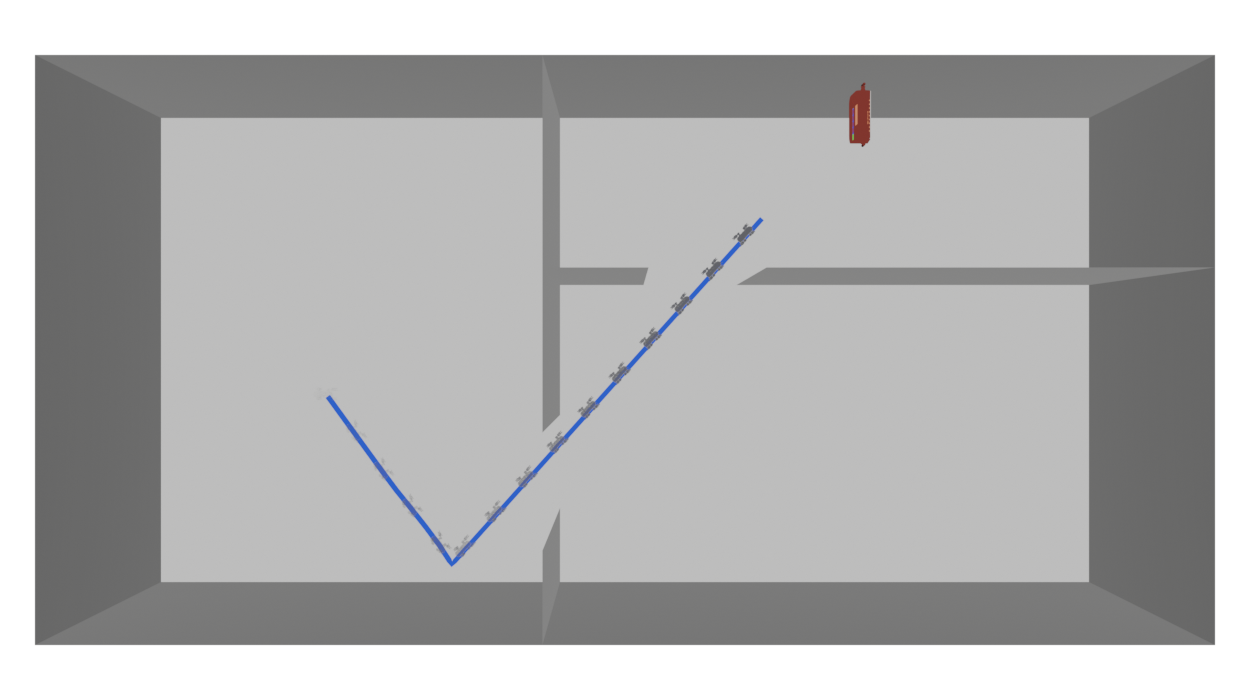}
        \label{fig: robot_frames}
    \end{minipage}
}
\subfigure[Frames (target perspective)]{
    \begin{minipage}[t]{0.22\textwidth}
     \includegraphics[width=\linewidth]{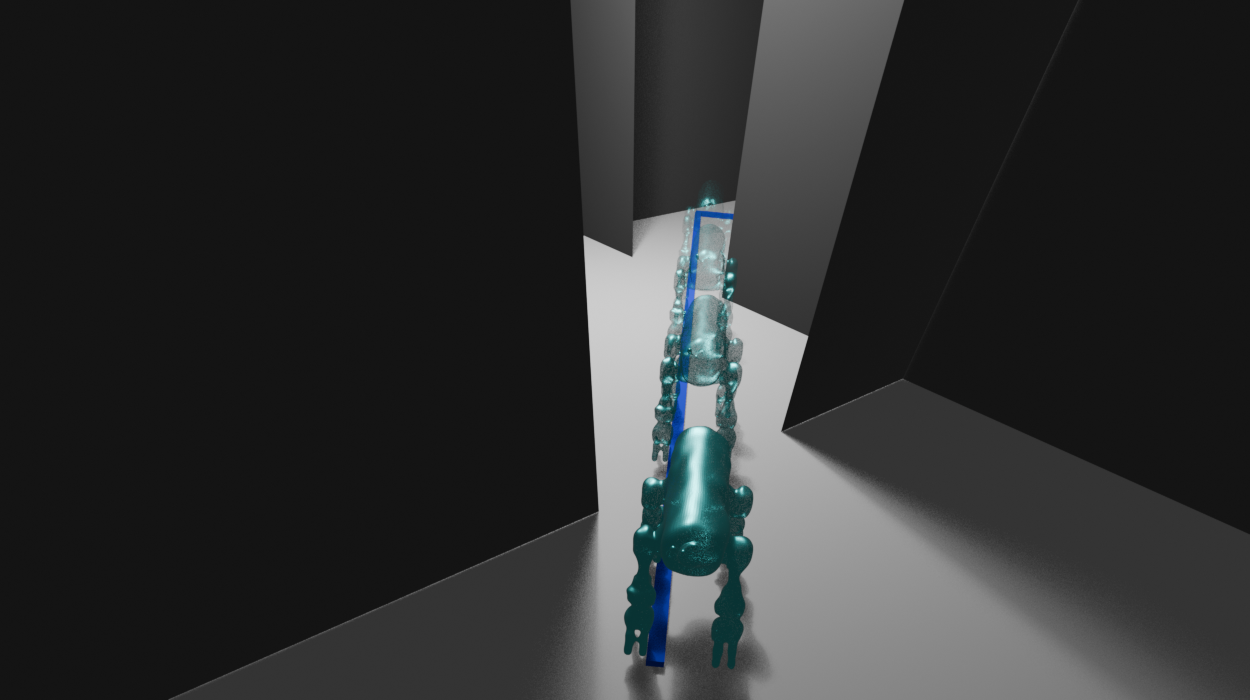}
        \label{fig: robot_frames at target perspective}
    \end{minipage}
}
\caption{
\ZP{Wave-Guided Robotic Navigation via Photon Splatting (Section IV.D): A robot equipped with a 15-element receiver array infers its trajectory solely from predicted multipath directions. Darker arrows indicate wave paths with the shortest time-of-delay. Comparison with ray tracing validates that Photon Splatting accurately guides the robot from a different room toward the transmitter using only predicted multipath components.}}

\label{fig: wi3rooms CIR}
\end{figure}

\ZP{This experiment evaluates Photon Splatting in a multi-room indoor environment, focusing on its ability to model directional multipath behavior and support wave-guided trajectory planning for mobile robots. Compared to previous datasets, the Wi3Rooms layout introduces structural complexity and signal occlusion, with multiple walls, narrow doorways, and strong multipath effects.}


\subsubsection{Wave-Guided Robotic Navigation}

\ZP{
Beyond conventional channel modeling tasks, this experiment demonstrates a novel downstream application of Photon Splatting: wave-informed motion planning. Specifically, unlike vision-based or geometry-based methods \cite{RoboticsFOCI2025, RoboticsGaussNav2025, RoboticsGSM2024, RoboticsGsPlanner2024, RoboticsSplatNav2025}, we explore how directional delay information can guide a mobile robot through obstructed environments toward a designated target.

The Wi3Rooms environment is drawn from the RPLAN dataset \cite{RPLAN2019} and consists of three interconnected rooms of varying size. As shown in Fig.~\ref{subfig: wi3rooms Tx/Rx planning (train dataset)}, the training dataset includes 400 transmitter locations and 3,750 receiver positions uniformly distributed across the floor plan. It is noted that to emphasize reflection and diffraction effects, all walls are modeled as non-penetrable, assuming no signal penetration. It is a reasonable approximation at 2.14 GHz for interior partitions with metal framing or high-reflectivity materials.
}


\subsubsection{Algorithms and Experiment}

\ZP{
To demonstrate this downstream application, we simulate autonomous navigation guided purely by wireless channel predictions. The task is to plan a trajectory from a robot's starting location to a fixed transmitter location, without access to the floor plan, visual input, or ray-tracing data. Instead, the robot relies on the directionality and delay information inferred from its predicted channel impulse responses 



The robot is equipped with a receiver array (see Fig.~\ref{fig: wi3rooms CIR}), which allows it to sample the CIR at multiple spatial offsets. At each time step, the robot queries Photon Splatting to obtain the CIR at each receiver in the array. It then identifies the strongest propagation direction determined by the shortest time-of-delay and moves one step forward in that direction. 
Based on this direction, the robot moves a small step forward, performs occlusion-aware path correction, and repeats the process. This loop continues until it reaches the vicinity of the transmitter, as summarized in Algorithm~\ref{algorithm: robotics}. It is noted that the robot has no prior access to the floor plan or ray geometry. It is guided purely by the learned multipath behavior embedded in photon splatting.

Fig.~\ref{fig: wi3rooms CIR} illustrates this process over three frames. Initially, the robot is located in a different room from the transmitter and has no direct line-of-sight. 
Yet, Photon Splatting accurately predicts multipath signals arriving via reflection and diffraction. As the robot discovers a door and moves closer to the transmitter, it dynamically updates its heading based on the inferred wavefronts, effectively tracing a physically valid path without explicit access to floorplan geometry or ray-tracing information.

This experiment highlights a unique advantage of Photon Splatting: the ability to infer navigational cues from learned wave interactions. It opens up new directions in wave-guided mobility and robotics, enabling real-time navigation, localization, and control based on channel physics alone. Such capability is especially valuable in RF-constrained or visually degraded environments where traditional sensors may fail.
}

\ZVP{\begin{algorithm}
\caption{\textbf{Wave-Informed Trajectory Planning Using Photon Splatting}}
\begin{algorithmic}[1]
\State Initialize \textbf{Robot.Rx.pos} $\gets (x_0, y_0, z_0)$
\State Initialize \textbf{Target.Tx.pos} $\gets (x_t, y_t, z_t)$
\State Initialize unit velocity $u_{m} \gets$ Input
\While{$\|\text{Robot.Rx.pos} - \text{Target.Tx.pos}\| > \epsilon$}
  \State $\text{CIR} \gets$ \texttt{PhotonSplatting(Scene, Tx, Rx)}
  \State Compute delay-based weights: $w_i \gets 1 / (c \cdot \tau_i)$
  \State Determine best direction: $\vec{d} \gets \arg\max_i(w_i)$
  \State Compute max allowable distance: \\ \qquad \quad $d_m \gets$ \texttt{SplatOcclusion(Rx, $\vec{d}$, Scene)}
  \State Compute desired movement: $d_u \gets \Delta t \cdot u_{m}$
  \If{$d_m > d_u$}
    \State $\text{walk} \gets d_u \cdot \vec{d}$
  \Else
    \State $\vec{n} \gets$ surface normal at occlusion point
    \State $\text{walk} \gets d_m \cdot \vec{d} + (d_u - d_m) \cdot \text{Reflect}(\vec{d}, \vec{n})$
  \EndIf
  \State \textbf{Robot.Rx.pos} $\gets$ \textbf{Robot.Rx.pos} $+ \text{walk}$
\EndWhile
\end{algorithmic}
\label{algorithm: robotics}
\end{algorithm}
}

\section{Conclusion and Limitation}

\ZVP{This paper introduces Photon Splatting, a real-time, physically informed neural framework for wireless channel modeling. Compared to existing literature, Photon Splatting predicts the full channel impulse response, including amplitude, delay, and angular profile, by learning wave interactions through photon-based environment representation and spherical harmonic encoding. We validate the framework across diverse environments: from a single-building outdoor scenario and cluttered indoor layouts to photorealistic caf\'e scenes. These experiments confirm the model's ability to generalize to unseen transmitter and receiver configurations, achieve real-time throughput across thousands of users, and preserve high physical fidelity. 

Beyond accuracy, we demonstrate downstream applications in multiuser MIMO precoding, antenna placement, and robotics trajectory planning. These capabilities make Photon Splatting a strong candidate for integration into emerging wireless digital twin systems, where runtime constraints, spatial awareness, and physical fidelity are critical.

Despite these strengths, the current pipeline assumes static scenes and known transmitter/receiver positions. Future work will focus on extending the framework to dynamic environments, enabling online updates from vision or LiDAR input, and jointly optimizing antenna placement or trajectory in a closed-loop fashion. With these extensions, we envision that Photon Splatting offers a scalable, explainable bridge between Maxwellian physics and real-time machine intelligence, which combines interpretability, generalization, and efficiency for the next generation of wireless systems.
}




\bibliographystyle{IEEEtran}
\bibliography{main}

\begin{thebibliography}{10}
\providecommand{\url}[1]{#1}
\csname url@samestyle\endcsname
\providecommand{\newblock}{\relax}
\providecommand{\bibinfo}[2]{#2}
\providecommand{\BIBentrySTDinterwordspacing}{\spaceskip=0pt\relax}
\providecommand{\BIBentryALTinterwordstretchfactor}{4}
\providecommand{\BIBentryALTinterwordspacing}{\spaceskip=\fontdimen2\font plus
\BIBentryALTinterwordstretchfactor\fontdimen3\font minus \fontdimen4\font\relax}
\providecommand{\BIBforeignlanguage}[2]{{%
\expandafter\ifx\csname l@#1\endcsname\relax
\typeout{** WARNING: IEEEtran.bst: No hyphenation pattern has been}%
\typeout{** loaded for the language `#1'. Using the pattern for}%
\typeout{** the default language instead.}%
\else
\language=\csname l@#1\endcsname
\fi
#2}}
\providecommand{\BIBdecl}{\relax}
\BIBdecl

\bibitem{DigitalTwin2022}
L.~Khan, W.~Saad, D.~Niyato, Z.~Han, and C.~Hong, ``\BIBforeignlanguage{English}{Digital-twin-enabled 6g: Vision, architectural trends, and future directions},'' \emph{\BIBforeignlanguage{English}{IEEE Communications Magazine}}, vol.~60, no.~1, pp. 74--80, Jan. 2022, publisher Copyright: {\textcopyright} 1979-2012 IEEE.

\bibitem{RadioRayTracing}
Z.~Yun and M.~F. Iskander, ``Ray tracing for radio propagation modeling: Principles and applications,'' \emph{IEEE access}, vol.~3, pp. 1089--1100, 2015.

\bibitem{Ray-Launching-Neural-2014}
L.~Azpilicueta, M.~Rawat, K.~Rawat, F.~M. Ghannouchi, and F.~Falcone, ``A ray launching-neural network approach for radio wave propagation analysis in complex indoor environments,'' \emph{IEEE Transactions on Antennas and Propagation}, vol.~62, no.~5, pp. 2777--2786, 2014.

\bibitem{8740286}
T.~Imai, K.~Kitao, and M.~Inomata, ``Radio propagation prediction model using convolutional neural networks by deep learning,'' in \emph{2019 13th European Conference on Antennas and Propagation (EuCAP)}, 2019, pp. 1--5.

\bibitem{DeepRay2022}
S.~Bakirtzis, K.~Qiu, J.~Zhang, and I.~Wassell, ``Deepray: Deep learning meets ray-tracing,'' in \emph{2022 16th European Conference on Antennas and Propagation (EuCAP)}, 2022, pp. 1--5.

\bibitem{Costas_2022}
A.~Seretis and C.~Sarris, ``An overview of machine learning techniques for radiowave propagation modeling,'' \emph{IEEE Transactions on Antennas and Propagation}, vol.~70, no.~6, pp. 3970--3985, 2022.

\bibitem{Seretis_2022_CNN}
A.~Seretis and C.~D. Sarris, ``Toward physics-based generalizable convolutional neural network models for indoor propagation,'' \emph{IEEE Transactions on Antennas and Propagation}, vol.~70, no.~6, pp. 4112--4126, 2022.

\bibitem{Seretis_2023}
A.~Seretis, C.~Xu, and C.~Sarris, ``Fast selection of indoor wireless transmitter locations with generalizable neural network propagation models,'' \emph{IEEE Transactions on Antennas and Propagation}, vol.~72, no.~10, pp. 7927--7940, 2024.

\bibitem{Liu2023_GNN}
S.~Liu, T.~Onishi, M.~Taki, and S.~Watanabe, ``A generalizable indoor propagation model based on graph neural networks,'' \emph{IEEE Transactions on Antennas and Propagation}, vol.~71, no.~7, pp. 6098--6110, 2023.

\bibitem{lee2023pmnet}
J.-H. Lee, O.~G. Serbetci, D.~P. Selvam, and A.~F. Molisch, ``Pmnet: Robust pathloss map prediction via supervised learning,'' in \emph{GLOBECOM 2023-2023 IEEE Global Communications Conference}.\hskip 1em plus 0.5em minus 0.4em\relax IEEE, 2023, pp. 4601--4606.

\bibitem{tse2005fundamentals}
D.~Tse, ``Fundamentals of wireless communication,'' \emph{Cambridge University Press google schola}, vol.~2, pp. 281--302, 2005.

\bibitem{FNO2020}
Z.~Li, N.~Kovachki, K.~Azizzadenesheli, B.~Liu, K.~Bhattacharya, A.~Stuart, and A.~Anandkumar, ``Fourier neural operator for parametric partial differential equations,'' \emph{arXiv preprint arXiv:2010.08895}, 2020.

\bibitem{865237}
C.~Brennan, P.~J. Cullen, and L.~Rossi, ``An {MFIE}-based tabulated interaction method for {UHF} terrain propagation problems,'' \emph{IEEE Transactions on Antennas and Propagation}, vol.~48, no.~6, pp. 1003--1005, Jun 2000.

\bibitem{Tsang_2004}
P.~Xu and L.~Tsang, ``Propagation over terrain and urban environment using the multilevel {UV} method and a hybrid {UV}/{SDFMM} method,'' \emph{IEEE Antennas and Wireless Propagation Letters}, vol.~3, pp. 336--339, 2004.

\bibitem{1504967}
C.~A. Tunc, A.~Altintas, and V.~B. Erturk, ``Examination of existent propagation models over large inhomogeneous terrain profiles using fast integral equation solution,'' \emph{IEEE Transactions on Antennas and Propagation}, vol.~53, no.~9, pp. 3080--3083, Sept 2005.

\bibitem{Sevgi_2007_Hybrid}
F.~Akleman and L.~Sevgi, ``A novel {MoM}- and {SSPE}-based groundwave-propagation field-strength prediction simulator,'' \emph{IEEE Antennas and Propagation Magazine}, vol.~49, no.~5, pp. 69--82, Oct 2007.

\bibitem{Sarris_FDTD_Fading}
A.~Alighanbari and C.~D. Sarris, ``Rigorous and efficient time-domain modeling of electromagnetic wave propagation and fading statistics in indoor wireless channels,'' \emph{IEEE Transactions on Antennas and Propagation}, vol.~55, no.~8, pp. 2373--2381, Aug 2007.

\bibitem{8485766}
B.~MacKie-Mason, Y.~Shao, A.~Greenwood, and Z.~Peng, ``Supercomputing-enabled first-principles analysis of radio wave propagation in urban environments,'' \emph{IEEE Transactions on Antennas and Propagation}, vol.~66, no.~12, pp. 6606--6617, 2018.

\bibitem{Aguado_2000}
F.~Aguado~Agelet, A.~Formella, J.~Hernando~Rabanos, F.~Isasi~de Vicente, and F.~Perez~Fontan, ``Efficient ray-tracing acceleration techniques for radio propagation modeling,'' \emph{IEEE Transactions on Vehicular Technology}, vol.~49, no.~6, pp. 2089--2104, 2000.

\bibitem{Sarkar_Ray_2001}
Z.~Ji, B.-H. Li, H.-X. Wang, H.-Y. Chen, and T.~K. Sarkar, ``Efficient ray-tracing methods for propagation prediction for indoor wireless communications,'' \emph{IEEE Antennas and Propagation Magazine}, vol.~43, no.~2, pp. 41--49, April 2001.

\bibitem{901882}
F.~S. de~Adana, O.~G. Blanco, I.~G. Diego, J.~P. Arriaga, and M.~F. Catedra, ``Propagation model based on ray tracing for the design of personal communication systems in indoor environments,'' \emph{IEEE Transactions on Vehicular Technology}, vol.~49, no.~6, pp. 2105--2112, Nov 2000.

\bibitem{Yun2015}
Z.~Yun and M.~F. Iskander, ``Ray tracing for radio propagation modeling: Principles and applications,'' \emph{IEEE Access}, vol.~3, pp. 1089--1100, 2015.

\bibitem{Raytracing_tutorial_2019}
D.~He, B.~Ai, K.~Guan, L.~Wang, Z.~Zhong, and T.~K{\"u}rner, ``The design and applications of high-performance ray-tracing simulation platform for 5{G} and beyond wireless communications: A tutorial,'' \emph{IEEE Communications Surveys \& Tutorials}, vol.~21, no.~1, pp. 10--27, 2019.

\bibitem{RenderEquation1986}
\BIBentryALTinterwordspacing
J.~T. Kajiya, ``The rendering equation,'' \emph{SIGGRAPH Comput. Graph.}, vol.~20, no.~4, p. 143–150, Aug. 1986. [Online]. Available: \url{https://doi.org/10.1145/15886.15902}
\BIBentrySTDinterwordspacing

\bibitem{gdpt2015}
\BIBentryALTinterwordspacing
M.~Kettunen, M.~Manzi, M.~Aittala, J.~Lehtinen, F.~Durand, and M.~Zwicker, ``Gradient-domain path tracing,'' \emph{ACM Trans. Graph.}, vol.~34, no.~4, Jul. 2015. [Online]. Available: \url{https://doi.org/10.1145/2766997}
\BIBentrySTDinterwordspacing

\bibitem{muller2017practical}
T.~M{\"u}ller, M.~Gross, and J.~Nov{\'a}k, ``Practical path guiding for efficient light-transport simulation,'' in \emph{Computer Graphics Forum}, vol.~36, no.~4.\hskip 1em plus 0.5em minus 0.4em\relax Wiley Online Library, 2017, pp. 91--100.

\bibitem{PhotonMapping1996}
H.~W. Jensen, ``Global illumination using photon maps,'' in \emph{Rendering Techniques’ 96: Proceedings of the Eurographics Workshop in Porto, Portugal, June 17--19, 1996 7}.\hskip 1em plus 0.5em minus 0.4em\relax Springer, 1996, pp. 21--30.

\bibitem{ProgressivePhotonMapping2008}
T.~Hachisuka, S.~Ogaki, and H.~W. Jensen, ``Progressive photon mapping,'' in \emph{ACM SIGGRAPH Asia 2008 papers}, 2008, pp. 1--8.

\bibitem{StochasticProgressivePhotonMapping2009}
T.~Hachisuka and H.~W. Jensen, ``Stochastic progressive photon mapping,'' in \emph{ACM SIGGRAPH Asia 2009 papers}, 2009, pp. 1--8.

\bibitem{NeRF}
B.~Mildenhall, P.~P. Srinivasan, M.~Tancik, J.~T. Barron, R.~Ramamoorthi, and R.~Ng, ``{NeRF}: representing scenes as neural radiance fields for view synthesis,'' \emph{Commun. ACM}, vol.~65, no.~1, p. 99–106, Dec. 2021.

\bibitem{Renerf2023}
Y.~Xu, G.~Zoss, P.~Chandran, M.~Gross, D.~Bradley, and P.~Gotardo, ``Renerf: Relightable neural radiance fields with nearfield lighting,'' in \emph{Proceedings of the IEEE/CVF International Conference on Computer Vision (ICCV)}, October 2023, pp. 22\,581--22\,591.

\bibitem{EyeNeRF2022}
\BIBentryALTinterwordspacing
G.~Li, A.~Meka, F.~Mueller, M.~C. Buehler, O.~Hilliges, and T.~Beeler, ``Eyenerf: A hybrid representation for photorealistic synthesis, animation and relighting of human eyes,'' \emph{ACM Trans. Graph.}, vol.~41, no.~4, jul 2022. [Online]. Available: \url{https://doi.org/10.1145/3528223.3530130}
\BIBentrySTDinterwordspacing

\bibitem{3DGS}
\BIBentryALTinterwordspacing
B.~Kerbl, G.~Kopanas, T.~Leimk{\"u}hler, and G.~Drettakis, ``3d gaussian splatting for real-time radiance field rendering,'' \emph{ACM Transactions on Graphics}, vol.~42, no.~4, July 2023. [Online]. Available: \url{https://repo-sam.inria.fr/fungraph/3d-gaussian-splatting/}
\BIBentrySTDinterwordspacing

\bibitem{2DGS}
B.~Huang, Z.~Yu, A.~Chen, A.~Geiger, and S.~Gao, ``2d gaussian splatting for geometrically accurate radiance fields,'' in \emph{SIGGRAPH 2024 Conference Papers}.\hskip 1em plus 0.5em minus 0.4em\relax Association for Computing Machinery, 2024.

\bibitem{fang2024mini}
G.~Fang and B.~Wang, ``Mini-splatting: Representing scenes with a constrained number of gaussians,'' in \emph{European Conference on Computer Vision}.\hskip 1em plus 0.5em minus 0.4em\relax Springer, 2024, pp. 165--181.

\bibitem{WiNert2023}
T.~Orekondy, K.~Pratik, S.~Kadambi, H.~Ye, J.~Soriaga, and A.~Behboodi, ``Winert: Towards neural ray tracing for wireless channel modelling and differentiable simulations,'' in \emph{The Eleventh International Conference on Learning Representations, {ICLR} 2023, Kigali, Rwanda, May 1-5, 2023}.\hskip 1em plus 0.5em minus 0.4em\relax OpenReview.net, 2023.

\bibitem{raypronet}
G.~Cao and Z.~Peng, ``Raypronet: A neural point field framework for radio propagation modeling in 3d environments,'' \emph{IEEE Journal on Multiscale and Multiphysics Computational Techniques}, vol.~9, pp. 330--340, 2024.

\bibitem{RF3DGS2024}
L.~Zhang, H.~Sun, S.~Berweger, C.~Gentile, and R.~Q. Hu, ``{RF-3DGS}: Wireless channel modeling with radio radiance field and 3d gaussian splatting,'' \emph{arXiv preprint arXiv:2411.19420}, 2024.

\bibitem{WRFGS2024}
C.~Wen, J.~Tong, Y.~Hu, Z.~Lin, and J.~Zhang, ``{WRF-GS}: Wireless radiation field reconstruction with 3d gaussian splatting,'' in \emph{IEEE INFOCOM 2025-IEEE Conference on Computer Communications}.\hskip 1em plus 0.5em minus 0.4em\relax IEEE, 2025, pp. 1--10.

\bibitem{aoudia2025sionna}
F.~A. Aoudia, J.~Hoydis, M.~Nimier-David, S.~Cammerer, and A.~Keller, ``Sionna {RT}: Technical report,'' \emph{arXiv preprint arXiv:2504.21719}, 2025.

\bibitem{sionna}
J.~Hoydis, S.~Cammerer, F.~{Ait Aoudia}, A.~Vem, N.~Binder, G.~Marcus, and A.~Keller, ``Sionna: An open-source library for next-generation physical layer research,'' \emph{arXiv preprint}, Mar. 2022.

\bibitem{RPLAN2019}
W.~Wu, X.-M. Fu, R.~Tang, Y.~Wang, Y.-H. Qi, and L.~Liu, ``Data-driven interior plan generation for residential buildings,'' \emph{ACM Transactions on Graphics (TOG)}, vol.~38, no.~6, pp. 1--12, 2019.

\bibitem{PyTorch}
\BIBentryALTinterwordspacing
J.~Ansel, E.~Yang, H.~He, N.~Gimelshein, A.~Jain, M.~Voznesensky, B.~Bao, P.~Bell, D.~Berard, E.~Burovski, G.~Chauhan, A.~Chourdia, W.~Constable, A.~Desmaison, Z.~DeVito, E.~Ellison, W.~Feng, J.~Gong, M.~Gschwind, B.~Hirsh, S.~Huang, K.~Kalambarkar, L.~Kirsch, M.~Lazos, M.~Lezcano, Y.~Liang, J.~Liang, Y.~Lu, C.~Luk, B.~Maher, Y.~Pan, C.~Puhrsch, M.~Reso, M.~Saroufim, M.~Y. Siraichi, H.~Suk, M.~Suo, P.~Tillet, E.~Wang, X.~Wang, W.~Wen, S.~Zhang, X.~Zhao, K.~Zhou, R.~Zou, A.~Mathews, G.~Chanan, P.~Wu, and S.~Chintala, ``{PyTorch 2: Faster Machine Learning Through Dynamic Python Bytecode Transformation and Graph Compilation},'' in \emph{29th ACM International Conference on Architectural Support for Programming Languages and Operating Systems, Volume 2 (ASPLOS '24)}.\hskip 1em plus 0.5em minus 0.4em\relax ACM, Apr. 2024. [Online]. Available: \url{https://pytorch.org/assets/pytorch2-2.pdf}
\BIBentrySTDinterwordspacing

\bibitem{hu2019taichi}
Y.~Hu, T.-M. Li, L.~Anderson, J.~Ragan-Kelley, and F.~Durand, ``Taichi: a language for high-performance computation on spatially sparse data structures,'' \emph{ACM Transactions on Graphics (TOG)}, vol.~38, no.~6, p. 201, 2019.

\bibitem{ravi2020pytorch3d}
N.~Ravi, J.~Reizenstein, D.~Novotny, T.~Gordon, W.-Y. Lo, J.~Johnson, and G.~Gkioxari, ``Accelerating 3d deep learning with pytorch3d,'' \emph{arXiv:2007.08501}, 2020.

\bibitem{ORCAAmazonBistro}
\BIBentryALTinterwordspacing
A.~Lumberyard, ``Amazon lumberyard bistro, open research content archive (orca),'' July 2017. [Online]. Available: \url{http://developer.nvidia.com/orca/amazon-lumberyard-bistro}
\BIBentrySTDinterwordspacing

\bibitem{RoboticsFOCI2025}
\BIBentryALTinterwordspacing
M.~G. Andreu, M.~Wilder-Smith, V.~Klemm, V.~Patil, J.~Tordesillas, and M.~Hutter, ``Foci: Trajectory optimization on gaussian splats,'' 2025. [Online]. Available: \url{https://arxiv.org/abs/2505.08510}
\BIBentrySTDinterwordspacing

\bibitem{RoboticsGaussNav2025}
\BIBentryALTinterwordspacing
X.~Lei, M.~Wang, W.~Zhou, and H.~Li, ``Gaussnav: Gaussian splatting for visual navigation,'' 2025. [Online]. Available: \url{https://arxiv.org/abs/2403.11625}
\BIBentrySTDinterwordspacing

\bibitem{RoboticsGSM2024}
\BIBentryALTinterwordspacing
K.~Goel and W.~Tabib, ``Distance and collision probability estimation from gaussian surface models,'' 2024. [Online]. Available: \url{https://arxiv.org/abs/2402.00186}
\BIBentrySTDinterwordspacing

\bibitem{RoboticsGsPlanner2024}
R.~Jin, Y.~Gao, Y.~Wang, Y.~Wu, H.~Lu, C.~Xu, and F.~Gao, ``Gs-planner: A gaussian-splatting-based planning framework for active high-fidelity reconstruction,'' in \emph{2024 IEEE/RSJ International Conference on Intelligent Robots and Systems (IROS)}.\hskip 1em plus 0.5em minus 0.4em\relax IEEE, 2024, pp. 11\,202--11\,209.

\bibitem{RoboticsSplatNav2025}
\BIBentryALTinterwordspacing
T.~Chen, O.~Shorinwa, J.~Bruno, A.~Swann, J.~Yu, W.~Zeng, K.~Nagami, P.~Dames, and M.~Schwager, ``Splat-nav: Safe real-time robot navigation in gaussian splatting maps,'' 2025. [Online]. Available: \url{https://arxiv.org/abs/2403.02751}
\BIBentrySTDinterwordspacing

\end{thebibliography}

\end{document}